\documentclass[journal]{IEEEtran}
\usepackage{amsmath,amsfonts}
\usepackage{algorithmic}
\usepackage{algorithm}
\usepackage{array}
\usepackage[caption=false,font=normalsize,labelfont=sf,textfont=sf]{subfig}
\usepackage{newtxtext}
\usepackage{textcomp}
\usepackage{stfloats}
\usepackage{url}
\usepackage{verbatim}
\usepackage{multirow}
\usepackage{graphicx}
\usepackage{cite}
\usepackage{hyperref}
\hypersetup{
    colorlinks=true,
    linkcolor=black,
    citecolor=black,
    urlcolor=blue,
}
\usepackage{xcolor}

\renewcommand{\color}[1]{}
\begin{document}

\title{When Simultaneous Localization and Mapping Meets \\ Wireless Communications: A Survey}

\author{Konstantinos Gounis, Sotiris A. Tegos,~\IEEEmembership{Senior Member,~IEEE}, Dimitrios Tyrovolas,~\IEEEmembership{Member,~IEEE} \\Panagiotis D. Diamantoulakis,~\IEEEmembership{Senior Member,~IEEE,} and George K. Karagiannidis,~\IEEEmembership{Fellow,~IEEE}
\thanks{(Corresponding author: George K. Karagiannidis)}
\thanks{
Konstantinos Gounis, Sotiris A. Tegos, Dimitrios Tyrovolas, Panagiotis Diamantoulakis, and George Karagiannidis are with the
Department of Electrical and Computer Engineering, Aristotle University of Thessaloniki, 54124 Thessaloniki, Greece
(e-mail: kgounis@ece.auth.gr; tegosoti@auth.gr; tyrovolas@auth.gr; padiaman@auth.gr; geokarag@auth.gr).}}
%

\markboth{IEEE COMMUNICATIONS SURVEYS AND TUTORIALS,~Vol.~XX, No.~XX, X~XXXX}%
{Shell \MakeLowercase{\textit{et al.}}: A Sample Article Using IEEEtran.cls for IEEE Journals}


\maketitle

\begin{abstract}
The availability of commercial wireless communication and sensing equipment combined with the advancements in intelligent autonomous systems {\color{blue}contribute to} robust joint communications and simultaneous localization and mapping (SLAM). Autonomous vehicles are one of the several applications that directly benefit from millimeter-wave (mmWave) access, {\color{blue}facilitating} improved communications and accurate localization in the interest of control and navigation. This paper surveys the state-of-the-art in the {\color{blue}intersection} of SLAM and Wireless Communications, attributing the bidirectional impact of each with a focus on visual SLAM (V-SLAM) integration. We provide an overview of key concepts related to {\color{blue}geometric channel modeling, and radio frequency (RF)-based localization and sensing.} {\color{blue}Several research directions are considered, including the interpretation of SLAM as a unified perception-estimation framework driven by wireless communications-based positioning and visual information, the integration of communication-aware objectives in active perception, and multi-modal odometry through the fusion of wireless measurements and complementary sensing modalities.} {\color{blue}We analyze estimation and control approaches such as Bayesian filters, feature-based pose estimation, perception-aware motion control, spatial methods for signal processing such as vector fields, and key technological aspects.} Among other interesting findings, we observe that monocular V-SLAM {\color{blue}might} greatly benefit from RF/mmWave {\color{blue}signals}, as the latter can serve as a proxy for scale ambiguity resolution. {\color{blue}Conversely, we find that wireless communications in the context of 5G and beyond can potentially benefit from visual odometry and landmark detection methods in the context of optimal geometric path prediction for wireless channels.} Moreover, we examine other perception sources besides {\color{blue}monocular vision} for SLAM such as Radio SLAM, wireless fidelity (WiFi) SLAM, and light detection and ranging (LiDAR) SLAM, and describe {\color{blue}a} twofold relation with wireless communications. Finally, integrated solutions performing joint communications and localization {\color{blue}appear to be} in their infancy: theoretical and practical advancements {\color{blue}could} add higher-level localization and semantic perception capabilities to RF/mmWave and multi-antenna technologies. 
\end{abstract}


\section{Introduction}
\IEEEPARstart{S}{IMULTANEOUS LOCALIZATION AND MAPPING} (SLAM) is a fundamental challenge in autonomous systems, requiring a robot to incrementally construct a map of an unknown environment while concurrently estimating its own position. Over the years, the robotics community has developed numerous successful approaches, leveraging diverse modeling techniques and sensor modalities to address this problem \cite{durrantwhyte2006slamreview}. To date, an increasing demand has been anticipated for improvements to SLAM regarding its knowledge discovery capacity and robust 3-D localization capabilities \cite{Li2024_RGBD_SLAM_YOLOv4MobileNetv3}.

In the last years, light has been shed on the capabilities of autonomous navigation systems such as the ability to understand, build maps in a metric range, and update them in a timely and memory-efficient manner, enabling real-time solutions to the original problem \cite{MurArtal2015_ORB_SLAM}. The enhancement of these capabilities is closely related to the advances made in computer vision during the last decade \cite{Evgenidis}. Hence, the research on V-SLAM is closely related to the scientific contributions recently made in the domain of deep learning (DL) \cite{Bruno2021_LIFT_SLAM}. Some notable examples are the DL-based extensions of the classical computer vision feature matching and tracking algorithms \cite{Alfarano2024_OpticalFlowReview, Ilg2017_FlowNet2}, or the robustification of V-SLAM with semantic segmentation for the removal of tracked features on objects that are more likely not to remain stationary \cite{Su2022_RealTimeDynamicSLAM}. The problem of SLAM has not been approached solely from the perspective of monocular / stereo vision. An alternative to the red-green-blue (RGB) image-based SLAM is the 3-D computer vision-based counterpart where LiDAR techniques have attracted the autonomous systems community’s interest due to their significantly accurate reconstruction capabilities \cite{He2024_LiDAR_SLAM_GNSS}. There are several notable studies that consider position estimation of an autonomous robot within a map that is concurrently built based on radio frequency (RF) and millimeter-wave (mmWave) propagation-based detections \cite{Yang2025_ISAC_Prototype, Du2024_GeneralSLAM_mmWave, Wei2024_rWiFiSLAM}.

On the other hand, wireless communications have undergone several impactful advancements, such as high-frequency and ultra-wide bandwidth capabilities, reduced latency, and increased coverage and reliability \cite{Shastri2022_mmWave_Localization_Review}. The advent of networks endowed with these capabilities has provided the physical foundation for this transformation, enabling wireless signals to convey both data and perceptual information. Among these, mmWave technologies, operating between 28 and 300 GHz, {\color{blue}have the potential} to support dense connectivity and precise spatial resolution \cite{Xiao2017_mmWaveEditorial_PartI}. As mmWave technology continues to evolve, it is deemed a {\color{blue} contributor} of sixth-generation (6G) mobile communication networks, which are expected to be commercially deployed around 2030. Driven by the need to support emerging applications such as connected autonomous vehicles, connected robots, and mixed reality, with all these applications exceeding the capabilities of current fifth-generation (5G) networks, researchers have outlined the 6G vision. 6G will aim at global coverage, considerably wide spectrum and strong security \cite{Zeng2024_EnvironmentAware_6G_Tutorial} as well as a single intelligent infrastructure where communication, sensing, learning, and control converge. Recently, the International Telecommunication Union (ITU) introduced a draft recommendation defining the framework and overarching objectives for International Mobile Telecommunications-2030 (IMT-2030) and beyond, highlighting six key usage scenarios: immersive communication, massive communication, hyper-reliable and low-latency communication (HRLLC), ubiquitous connectivity, and the integration of advanced artificial intelligence (AI) algorithms with communications \cite{Zhang2019_6G_Wireless_Vision}. 

Next-generation wireless communication technology serves communication purposes while also providing high-resolution user equipment (UE)-based localization and device-free sensing. These capabilities are inherent to the physics of mmWave propagation. To begin with, the shorter wavelength of mmWaves, compared to sub-6G signals, aids in low error-induced position estimation. Additionally, mmWave sensing {\color{blue}can result} in higher spatial scanning resolution due to propagation characteristics such as dominance of the line-of-sight (LOS) propagation component over the scattered ones that follow from propagation in areas with many reflecting surfaces \cite{Shastri2022_mmWave_Localization_Review}. Nevertheless, mmWave propagation {\color{blue}has the potential to achieve} long-range sensing, the latter at the cost of higher path loss and attenuation compared to lower frequency radio-waves. This can be tackled by: a) employing large or massive antenna arrays and b) devising specific protocols for initial access as well as beam training in the context of IEEE 802.11ay standard, in order for the electromagnetic field to be focused on a desired region in the cartesian space \cite{Zhou2018_IEEE80211ay_mmWaveWLANs}. 

Device-free localization (DFL), also known as radio sensing or sensor-less sensing, {\color{blue}can retrieve} object locations by analyzing wireless link measurements. The key wireless properties used in this technique are received signal strength (RSS) and channel state information (CSI). DFL methods fall into two main categories based on how they utilize the aforementioned measurements \cite{Shit2019_UbiLoc_Survey}. The first one, link quality measurement, assesses changes in signal strength caused by objects obstructing the wireless link. A probabilistic/stochastic model for the three-dimensional position state prior distribution can be combined with RSS measurement updates to estimate state posterior distribution, adopting a simplified odometry assumption such as a constant-velocity model. The second category, titled reflection or scattering of the link, relies on signal reflections and scattering from objects to estimate their location, making it analogous to a radar-based approach \cite{Shit2019_UbiLoc_Survey}. A typical scenario would include a number of transmitter-receiver pairs operating as a network of sensor radars, whilst in addition to the target objects, background objects are considered. Multi-object detection and localization can be established by placing nodes with ultra-wideband (UWB) sensing capabilities in the vicinity of the region of interest for target detection. The computation can be based on the time of arrival (ToA) of the back-scattered signals.

Although {\color{blue}several} recent surveys have investigated various aspects of wireless localization and sensing, few have examined the bidirectional nexus between SLAM or its core computer vision related elements and wireless communications. {\color{blue} Among the most closely related works, Katare et al.\cite{Katare2023_ApproximateEdgeAI_AutonomousDriving} reviewed the foundational elements of automated driving, assigning each to the related sense–think–act functionality as well as autonomous-driving perception, SLAM, and vehicle-to-everything (V2X) communications. However, their main focus was energy-efficient approximate edge AI, not the bidirectional SLAM–wireless estimation and control interplay.} Zeng et al. \cite{Zeng2024_EnvironmentAware_6G_Tutorial} provided a tutorial on environment-aware communications through the concept of channel knowledge maps (CKM), highlighting fingerprint-based localization and integrated sensing–communication frameworks. However, it remains limited to the cellular-centric extraction of positioning and spatial information without extending into autonomous SLAM or computer vision-aided wireless communication aspects. Italiano et al. \cite{Italiano2025_5G_Positioning_Tutorial} presented a comprehensive tutorial on 5G capabilities, detailing THz-band imaging and high-frequency SLAM for highly accurate positioning and NLOS scenarios leveraging multipath reflections. Moreover, they included in their work reconfigurable intelligent surfaces (RIS)-aided holographic localization and probabilistic SLAM-inspired data association, yet their treatment remains confined to UE localization accuracy and does not unify mathematical SLAM foundations or vision-driven autonomy. Similarly, Shastri et al. \cite{Shastri2022_mmWave_Localization_Review} and Cheraghinia et al. \cite{Cheraghinia2025_UWB_Radar_Overview} covered device-based and device-free mmWave and UWB radar localization, respectively, offering valuable insights into channel-level processing, angle-of-arrival estimation, and tracking. However, their work did not involve an integral perspective incorporating such modalities into broader SLAM frameworks involving multi-sensor fusion, control, and learning-based mapping.

Beyond these highly overlapping works, moderately related surveys such as Primeau et al. \cite{Primeau2018_CI_in_WSANS}, Moura et al. \cite{Moura2019_GameTheory_MEC_Survey}, and Kong et al. \cite{Kong2025_mmWaveRadar_Survey} examined computational intelligence-based wireless sensor and actuator networks, opportunistic localization for sensor networks, and mmWave Radar based sensing with a focus on machine learning (ML) techniques, respectively. However, each study addressed these mechanisms in an uncombined computer vision and wireless communications SLAM setting. Other reviews such as Liu et al. \cite{Liu2022_ISAC_FundamentalLimits} focused on integrated sensing and communication (ISAC) limits, highlight the rising interest in environment-aware communication, yet none have connected these to the algorithmic or autonomy-{\color{blue}supporting} layers of SLAM. In {\color{blue}comparison}, the present survey synthesizes these research threads into a unified {\color{blue}framework comprising} (i) the foundations of autonomous SLAM, and (ii) mathematical methods and multimodal SLAM techniques, while highlighting the transformative role of modern wireless communication technologies in the era of 6G, ranging from WiFi fingerprinting and radio map construction to deep neural network (DNN)–based wireless SLAM and RIS-aided MetaSLAM. This integrative treatment {\color{blue}can provide} a cross-disciplinary reference point for understanding how wireless signals are increasingly redefining localization, mapping, and situational awareness across next-generation robotic and communication systems. 

{\color{blue}
The literature search was conducted using a three-step approach to ensure broad coverage of relevant studies. First, a search was performed in the Scopus database using predefined keywords 
“(SLAM OR Simultaneous Localization and Mapping) AND (5G OR 6G OR WiFi OR Wireless Communications*). The search was limited to articles published up to September 2025. Second, a targeted manual search was conducted of all surveys published over the period January 2018 – September 2025 in IEEE Communications Surveys \& Tutorials, to capture recent surveys in the field. This step was undertaken to ensure inclusion of influential and field-specific syntheses that may not have been fully captured through the Scopus database indexing. Finally, reference lists of the identified articles/surveys were screened to identify additional relevant studies and reports not retrieved through the latter search.}

\par
{\color{blue}
Table~\ref{tab:survey_comparison_revised} compares this survey with closely related works in terms of the following dimensions: 
(i) SLAM as a core framework, 
(ii) joint consideration of vision - wireless communications, 
(iii) inclusion of mathematical models for estimation, and 
(iv) passive and (v) active SLAM within a combined perception - communications - control approach.

\begin{table*}[t]
\centering
{\color{blue}
\caption{Comparison of Representative Surveys and Tutorials Related to SLAM and Wireless Communications}
\label{tab:survey_comparison_revised}
\scriptsize
\setlength{\tabcolsep}{3.2pt}
\renewcommand{\arraystretch}{1.18}
\begin{tabular}{|p{2.9cm}|p{2.8cm}|c|c|c|c|c|}
\hline
\textbf{Survey} & \textbf{Primary Focus} & \textbf{SLAM as a Core Framework} & \textbf{Joint Vision-Wireless} & \textbf{Mathematical Models} & \textbf{Passive SLAM} & \textbf{Active SLAM} \\
\hline
Zhou et al.~\cite{Zhou2018_IEEE80211ay_mmWaveWLANs}
& IEEE 802.11ay mmWave WLAN design, beam training
& No & Low & No & No & No \\
\hline
del Peral-Rosado et al.~\cite{delperalrosado2018survey}
& Cellular mobile radio localization from 1G to 5G
& No & Low & No & Moderate & No \\
\hline
Primeau et al.~\cite{Primeau2018_CI_in_WSANS}
& Computational intelligence in wireless sensor and actuator networks
& No & Low & No & Yes & High \\
\hline
Moura and Hutchison~\cite{Moura2019_GameTheory_MEC_Survey}
& Game theory for multi-access edge computing, opportunistic localization
& No & Low & Moderate & No & No \\
\hline
Shit et al.~\cite{Shit2019_UbiLoc_Survey}
& Device-free localization taxonomy for smart environments
& No & Low & High & High & No \\
\hline
Liu et al.~\cite{Liu2022_ISAC_FundamentalLimits}
& Fundamental limits of integrated sensing and communication
& No & Low & Yes & Low & No \\
\hline
Shastri et al.~\cite{Shastri2022_mmWave_Localization_Review}
& mmWave device-based localization and device-free sensing
& No & Low & High & High & No \\
\hline
Zeng et al.~\cite{Zeng2024_EnvironmentAware_6G_Tutorial}
& Environment-aware communications via CKM for 6G, environment-aware localization and sensing with a focus on mathematical models for estimation
& No & Low & Yes & No & No \\
\hline
Katare et al.~\cite{Katare2023_ApproximateEdgeAI_AutonomousDriving}
& Energy-aware approximate edge AI, sense–think–act scheme in autonomous systems, perception, object detection, RGB / LiDAR visual SLAM, V2X communications
& Moderate & High & Low & Yes & Low \\
\hline
Italiano et al.~\cite{Italiano2025_5G_Positioning_Tutorial}
& 5G positioning, mathematical models attributing different wireless measurement signals (e.g. ToA, AoA), UE positioning, mapping in harsh NLOS
& Moderate & No & Yes & Low & No \\
\hline
Kong et al.~\cite{Kong2025_mmWaveRadar_Survey}
& mmWave radar-based sensing applications and signal processing, focus on localization.
& Moderate & No & Moderate & Low & No \\
\hline
Cheraghinia et al.~\cite{Cheraghinia2025_UWB_Radar_Overview}
& UWB radar applications, signal processing, datasets, and trends
& Moderate & Moderate & High & Yes & No \\
\hline
\textbf{This Survey}
& SLAM - wireless communications integration in autonomous systems
& \textbf{Yes} & \textbf{Yes} & \textbf{Yes} & \textbf{Yes} & \textbf{Yes} \\
\hline
\end{tabular}
}
\end{table*}

The comparison was based on a scoring methodology presented in Table~\ref{tab:scoring_logic}. 
Each dimension is assessed using four defined binary checks, which are assigned a value of 1 only when clear supporting evidence is identified in the corresponding paper, and 0 otherwise.
The total score per dimension was computed as the sum of its four checks, resulting in a state in $\{0/4, 1/4, 2/4, 3/4, 4/4\}$.

These states are mapped to qualitative levels as follows:
\textbf{No} ($0/4$), 
\textbf{Low} ($1/4$), 
\textbf{Moderate} ($2/4$), 
\textbf{High} ($3/4$), and 
\textbf{Yes} ($4/4$), 
reflecting increasing degrees of alignment with the defining characteristics of each dimension.
}

\begin{table*}[t]
\centering
{\color{blue}
\caption{Dimensions' Scoring Criteria and Assessment Logic for the Comparison Framework}
\label{tab:scoring_logic}
\scriptsize
\setlength{\tabcolsep}{3.2pt}
\renewcommand{\arraystretch}{1.18}

\begin{tabular}{|c|p{3.8cm}|c|p{5.8cm}|c|}
\hline
\textbf{Dimension} & \textbf{Criterion} & \textbf{Check} & \textbf{Binary Condition for Assignment} & \textbf{Contribution} \\
\hline

\multirow{4}{*}{SLAM as a Core Framework} 
& SLAM central in title / abstract / introduction 
& 1 
& Explicit mention of SLAM as a primary focus in title, abstract, or introduction 
& \multirow{4}{*}{0--4} \\

& Mapping jointly treated with localization 
& 2 
& Joint discussion of mapping and localization processes within the survey 
& \\

& SLAM components / pipelines discussed 
& 3 
& Significant treatment of SLAM pipelines, modules, or algorithmic components 
& \\

& SLAM as the organizing lens 
& 4 
& Survey structure is centered around SLAM concepts or taxonomy 
& \\
\hline

\multirow{4}{*}{Joint Vision - Wireless} 
& Vision modalities discussed 
& 1 
& Inclusion of RGB, stereo, RGB-D or LiDAR vision-based sensing modalities 
& \multirow{4}{*}{0--4} \\

& RF / mmWave communications discussed 
& 2 
& Discussion of 5G / 6G, RF / mmWave (cellular or WiFi) communication systems 
& \\

& Joint analytical treatment 
& 3 
& Vision and wireless systems are treated within a unified analytical / modeling framework 
& \\

& Explicit integration 
& 4 
& Direct discussion of interaction or fusion between vision and wireless modalities 
& \\
\hline

\multirow{4}{*}{Mathematical Models} 
& Measurement / signal / state-space models 
& 1 
& Presence of explicit equations, signal models, or state-space formulations 
& \multirow{4}{*}{0--4} \\

& Estimation/probabilistic methods 
& 2 
& Use of stochastic, Bayesian / probabilistic estimation techniques 
& \\

& Optimization/bounds 
& 3 
& Central use of optimization problems, bounds, or objective functions 
& \\

& Analytical depth 
& 4 
& Predominantly analytical treatment 
& \\
\hline

\multirow{4}{*}{Passive SLAM} 
& SLAM in perception - communications - control context 
& 1 
& Localization / mapping embedded in perception – communication – control systems 
& \multirow{4}{*}{0--4} \\

& Link to control/autonomy 
& 2 
& State estimation explicitly connected to control or decision-making 
& \\

& System-level coupling 
& 3 
& Integration across system layers rather than isolated modules 
& \\

& Passive formulation 
& 4 
& No explicit action optimization or active information-seeking mechanism required 
& \\
\hline

\multirow{4}{*}{Active SLAM} 
& Control / action discussed in perception – communication – control context 
& 1 
& Explicit treatment of control inputs or action selection mechanisms 
& \multirow{4}{*}{0--4} \\

& Action improves estimation / communication 
& 2 
& Actions linked to uncertainty reduction, mapping, localization, or communication performance 
& \\

& Active perception loop 
& 3 
& Closed-loop perception or information-seeking behavior is present 
& \\

& Action-estimation coupling 
& 4 
& Control integrated within estimation or sensing-aided communication loops 
& \\
\hline

\end{tabular}

\vspace{0.3cm}

\begin{minipage}{0.95\textwidth}
\footnotesize
\textit{Scoring Rule:} Each dimension is assessed using four binary checks. A check is assigned value 1 when supporting evidence is identified in the paper, otherwise, it is assigned conservatively 0. The total score per axis is computed as the sum of its four checks, yielding a state in $\{0/4, 1/4, 2/4, 3/4, 4/4\}$, which is mapped to the qualitative labels \textbf{No}, \textbf{Low}, \textbf{Moderate}, \textbf{High}, and \textbf{Yes}, respectively.
\end{minipage}
}
\end{table*}

Specifically, this survey provides a structured and in-depth investigation of the emerging intersection of SLAM and wireless communications, {\color{blue}organizing} both foundational and new developments. {\color{blue}Centered on} three core thematic axes: the foundations of SLAM in networked autonomous systems, SLAM methods and techniques, and the advancements of wireless communications in SLAM. In the aforementioned areas of focus, {\color{blue}an overview} of the state-of-the-art is provided. Our study {\color{blue}describes} the pillars for networked autonomous systems' SLAM, highlighting the critical role of the computer vision algorithms for extracting features that can serve as the basis for reconstruction of the environment as well as of the control laws in autonomous and connected systems. Based on these pillars, the paradigm of passive SLAM is highlighted within the larger picture of autonomous systems, where SLAM has the responsibility of being a state observer using diverse radio and visual features. Optimization of the control inputs in order for a networked autonomous system to reduce its ego-motion or landmark mapping uncertainty is considered afterwards, {\color{blue}highlighting} the domain of active estimation or active SLAM. The latter paradigm {\color{blue} could} be central to a pipeline for achieving communication goals, i.e., by conversion of the wireless communications' objective such as path-loss reduction/RSS improvement through optimal beam steering to a perceptual goal. Thus, a goal such as improving the accuracy of UE position and orientation through the execution of specific motion commands {\color{blue}could} be directly translated into a beam prediction objective. 

On the methodological {\color{blue}perspectives}, the survey focuses on mathematical formulations, RF and computer vision–based localization paradigms, LiDAR and inertial SLAM, and {\color{blue}algorithmic elements} {\color{blue} in} modern SLAM systems. Furthermore, the survey {\color{blue}examines} the integration of wireless technologies into SLAM pipelines, {\color{blue}addressing} developments in device-based sensing, WiFi fingerprinting-based localization, radio map construction via wireless sensing, and emerging DNN-driven wireless SLAM architectures. By organizing these contributions in a unified framework, the survey demonstrates how wireless signals augment the vehicle state vector and redefine localization and mapping strategies in an advantageous manner. As the wireless communications-computer vision intersection lives in the core of this study, effort is put into showcasing the importance of computer vision pipelines for predicting the environment state. This piece of information can be particularly useful for predicting future link quality. 

The organization of the paper is as follows. Section II {\color{blue}provides the historical background.} Section III presents the foundations of SLAM in networked autonomous systems. {\color{blue}SLAM methods and modality-based techniques are presented in Section IV. Section V discusses wireless- and 6G-enabled SLAM. Section VI discusses challenges and future directions.} Finally, Section VII concludes the paper.

\section{Historical Background}

\begin{figure}[!t]
\centering
\includegraphics[width=\linewidth]{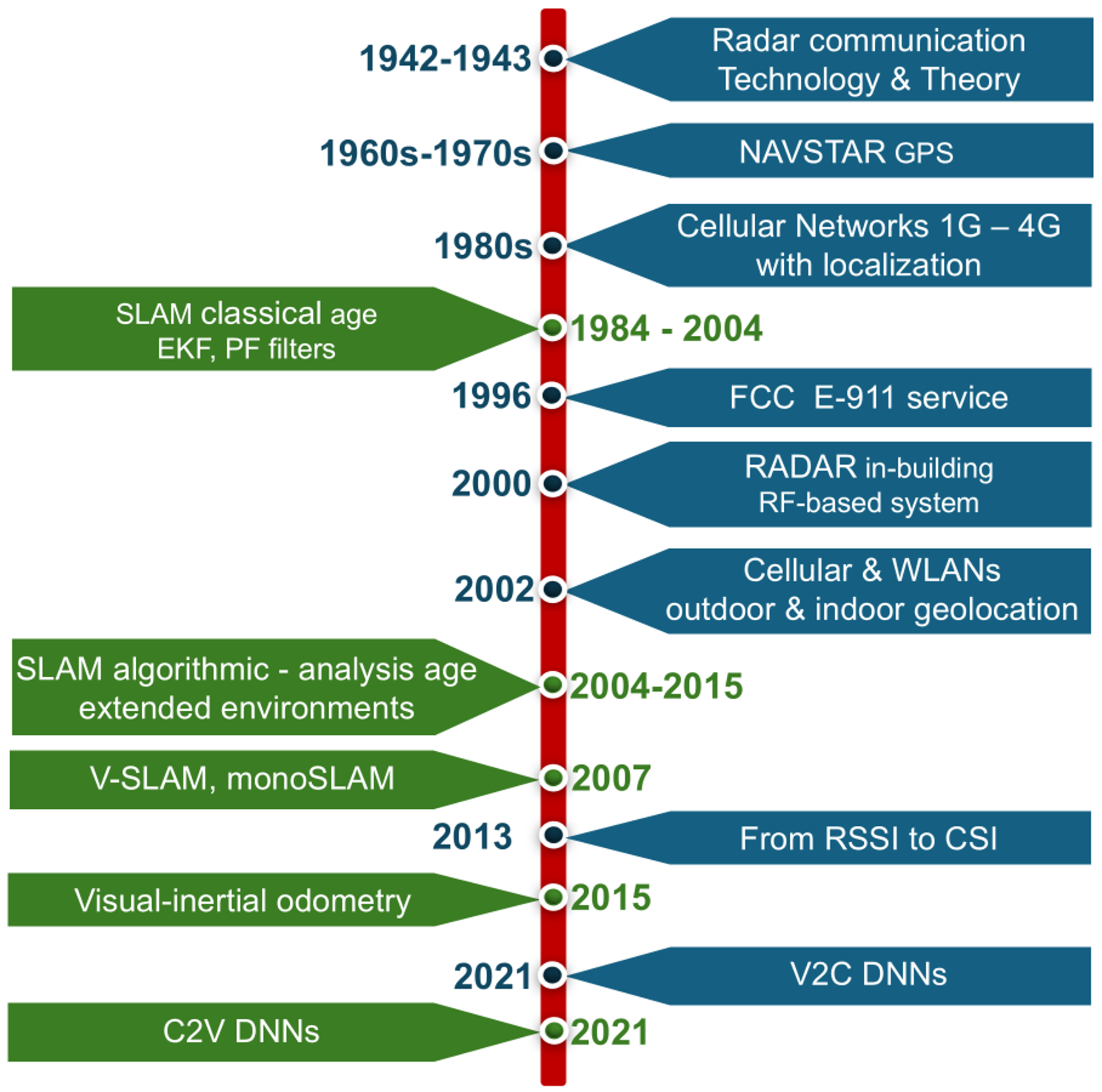}
\caption{{\color{blue}Timeline of significant developments in the intersection of SLAM and wireless communication technologies. Green indicates SLAM developments and blue the wireless communication technologies.}}
\label{fig_1}
\end{figure}

The intersection of SLAM with wireless communications represents the convergence of two engineering domains that have evolved independently yet with complementary objectives. Although SLAM originates from robotics and autonomous navigation, wireless communications have long provided the infrastructure for connectivity and RF-based sensing. Examining their historical trajectories in parallel highlights how advances in navigation, signal processing, and communication progressively enabled more integrated positioning and mapping capabilities. {\color{blue}Fig \ref{fig_1} shows a historical overview of significant developments in the intersection of SLAM and wireless communication technologies.} 

{\color{blue}\subsection{Wireless Communications and localization evolution}}
The origins of RF-based localization can be traced to the mid-20th century, when radar technology emerged as one of the foremost scientific developments of World War II \cite{bell1943radar}. Beyond its critical wartime role, radar introduced the principle of exploiting RF signals for detection and ranging, which became a cornerstone for subsequent navigation systems. At nearly the same time, Wiener established the theoretical foundations of modern communication theory \cite{wiener1949extrapolation}. His work on stochastic processes and the formulation of the Wiener filter provided rigorous tools for signal estimation in noisy environments, fundamentally shaping the design of communication systems. Together, these advances demonstrated both the feasibility of RF-based sensing and the mathematical means of optimizing its performance. This combination of technological and theoretical progress laid the groundwork for satellite navigation systems such as the global positioning system (GPS).

In the mid-1960s, the U.S. Air Force and Navy initiated the NAVSTAR GPS program \cite{heuerman1983gps}, which over subsequent decades evolved into a global, all-weather navigation system capable of serving both military and civilian applications. The success of GPS highlighted the potential of satellite-based positioning but also revealed its limitations in obstructed or indoor environments. In parallel, the rapid expansion of cellular networks during the 1980s and 1990s enabled the first network-assisted positioning systems. Although designed primarily to provide communication services, successive generations of cellular networks incorporated positioning functions as complementary features. This trend is well documented in \cite{delperalrosado2018survey}, which reviewed localization methods that were standardized from the first to the fourth generation (1G–4G) of cellular systems and outlined how positioning requirements became increasingly integral to mobile communications. By this stage, positioning was no longer viewed as an optional add-on but as a functional requirement tied directly to the evolution of communication infrastructures.

The proliferation of mobile devices and wireless local-area networks in the late 1990s and early 2000s further intensified interest in location-aware services, particularly in indoor environments where satellite-based positioning is severely degraded. A seminal contribution in this area was the RADAR system \cite{bahl2000radar}, which utilized RSS measurements from multiple access points to estimate user location within buildings. This line of work opened a broad research field in wireless indoor localization, employing diverse signal modalities such as acoustic signals, ultrasound, FM radio, infrared, Bluetooth, WiFi, ultra-wideband (UWB), and received signal strength indicator (RSSI) analysis \cite{yang2013rssi2csi}. The shortcomings of global navigation satellite systems (GNSS) in multipath-rich and obstructed scenarios provided strong motivation for such developments, with UWB in particular attracting attention due to its high time resolution, large bandwidth, low-cost implementation, and low spectral density relative to WiFi and Bluetooth. Nevertheless, challenges such as multipath propagation and non-line-of-sight (NLOS) conditions limited UWB performance, prompting novel approaches, including the deployment of mobile aerial sensors as anchor points to enhance coverage flexibility \cite{nosrati2022uwbdnn}. This trend was further reinforced by regulatory requirements, such as the Federal Communications Commission (FCC) mandate, which required wireless service providers to deliver accurate location information for Enhanced 911 (E-911) emergency services \cite{Sayed2005_WirelessLocation}. These requirements accelerated research on hybrid architectures combining cellular and wireless local area networks (WLANs) for both outdoor and indoor geolocation, as surveyed in \cite{Pahlavan2002_IndoorGeolocation}. Thus, by the turn of the century, wireless localization had progressed from military-driven positioning to consumer and regulatory-driven deployments, establishing a parallel trajectory to the robotics domain, where SLAM was beginning to emerge.

As wireless systems evolved, localization and communication began to be designed as interdependent functions. The evolution toward higher-frequency operation and programmable propagation environments introduced mmWave multiple-input, multiple-output (MIMO) systems enhanced with RIS, which were recognized for their ability to provide fine-grained control of the radio environment \cite{ganapathy2025anm}. At the same time, their practical deployment revealed persistent challenges related to channel estimation and phase-shift optimization, which motivated further refinements. Frameworks were subsequently explored in which orthogonal frequency division multiplexing (OFDM) coexisted with RIS, allowing location information derived from RIS-assisted communications to support more efficient user multiplexing strategies \cite{saggese2023localization}. During this stage, fingerprint-based localization methods also emerged as adaptations to dynamic NLOS conditions, demonstrating the potential to sustain robust operation in real-world MIMO-OFDM systems \cite{ismayilov2024domainadaptation}. These advances reflect the broader progression from standalone RF-based navigation tools to tightly integrated communication-localization frameworks, where accurate positioning is no longer an ancillary feature but an expected capability of modern cellular infrastructures. This system-level integration in wireless communications provides a natural counterpart to developments in robotics, where mapping and localization matured along a parallel trajectory.

{\color{blue}\subsection{SLAM evolution}}
In parallel with these developments in wireless communications, robotics was establishing the foundations of SLAM as a framework for autonomous navigation in unknown environments. Probabilistic formulations for representing spatial uncertainty were introduced in \cite{smith1986spatialuncertainty}, marking the beginning of SLAM as a formal estimation problem. Building on this foundation, the authors of \cite{leonard1991slam} formalized the SLAM problem in the 1990s, proving that convergence was achievable under suitable conditions and framing the challenge as the interdependence of trajectory estimation and map construction, often expressed as the question of which comes first, the map or the motion. As computational resources expanded and sensor technologies diversified, including cameras, laser scanners, and radar, the scope of SLAM broadened significantly \cite{bresson2017slamtrends}. Comprehensive reviews in \cite{durrantwhyte2006slamreview, bailey2006slam2} described this period as the “classical age” of SLAM (1986–2004), dominated by stochastic formulations such as extended Kalman filter (EKF)-based and particle filter (PF)-based methods \cite{thrun2005probabilistic}. To address the scalability challenges of feature-based approaches in large environments, refinements such as the Rao-Blackwellized particle filter (RBPF) were introduced, significantly improving the efficiency of grid mapping \cite{grisetti2007gridmapping}. These advances established the probabilistic foundations and scalable algorithms that framed the classical phase of SLAM, setting the stage for subsequent efforts aimed at greater robustness and adaptability.

{\color{blue}\subsection{Intersecting Trajectories}}
The next stage in the evolution of SLAM emphasized robustness, scalability, and the integration of multiple sensing modalities. The review in \cite{cadena2016slamfuture} highlighted long-term mapping, semantic representations, and active exploration as emerging research priorities, reflecting the need for SLAM systems to perform reliably over extended deployments and in increasingly complex environments. During this period, V-SLAM gained prominence, with MonoSLAM \cite{davison2007monoslam} introducing real-time monocular trajectory estimation, while subsequent frameworks such as \cite{Leutenegger2015} demonstrated how coupling visual and inertial measurements through nonlinear optimization could further enhance accuracy and stability. These developments established visual–inertial odometry as a central approach in SLAM research. At the same time, the integration of machine learning marked another important stage, with \cite{Han2024} reviewing the transition toward data-driven methods and works such as \cite{Li2021_DeepSLAM, Zhang2025} illustrating how DL approaches were introduced to address the limitations of feature-based techniques, marking the beginning of a shift toward data-driven SLAM. In parallel, computer vision began to intersect with wireless communication, as demonstrated in \cite{Nishio2021}, where visual inputs were employed to predict mmWave channel blockages before they occurred, while related efforts proposed view-to-communication (V2C) and communication-to-view (C2V) paradigms that fused information from RGB-D, LiDAR, and radar sensing. Taken together, these advances marked a transition from purely geometric estimation toward systems capable of fusing multimodal data streams and incorporating learning-based representations.

The convergence of these approaches became increasingly apparent as SLAM concepts were adapted to wireless infrastructures and communication systems began to incorporate mapping capabilities. Positioning, which had once been considered auxiliary, became essential not only for autonomous robotics but also for mobile systems supporting public safety, transportation, and broadband services. Although traditional SLAM implementations using LiDAR or high-resolution cameras offered high accuracy, their cost and complexity motivated alternative approaches, including radar-based SLAM for handheld devices \cite{Lotti2023_RadioSLAM} and portable THz sensing prototypes \cite{Li2021_THZSensing}. Comparative evaluations, such as \cite{Yapar2023_RealTimeLocalization}, reported that convolutional neural networks could exceed the performance of conventional RSS and ToA-based methods in urban localization, illustrating how learning-based solutions were gradually integrated into both communities to overcome persistent limitations. Viewed historically, the progression of wireless localization and SLAM reveals common parallel patterns as they both began with probabilistic models, advanced through scalable algorithms and multimodal sensing, and are now converging toward ISAC systems. To this end, this convergence emerges as the cumulative result of decades of advances in navigation, perception, and communication, laying the groundwork for future systems where localization and connectivity are inseparably intertwined.

\section{Foundations of SLAM in Networked Autonomous Systems}

\begin{figure}[!t]
\centering
\includegraphics[width=\linewidth]{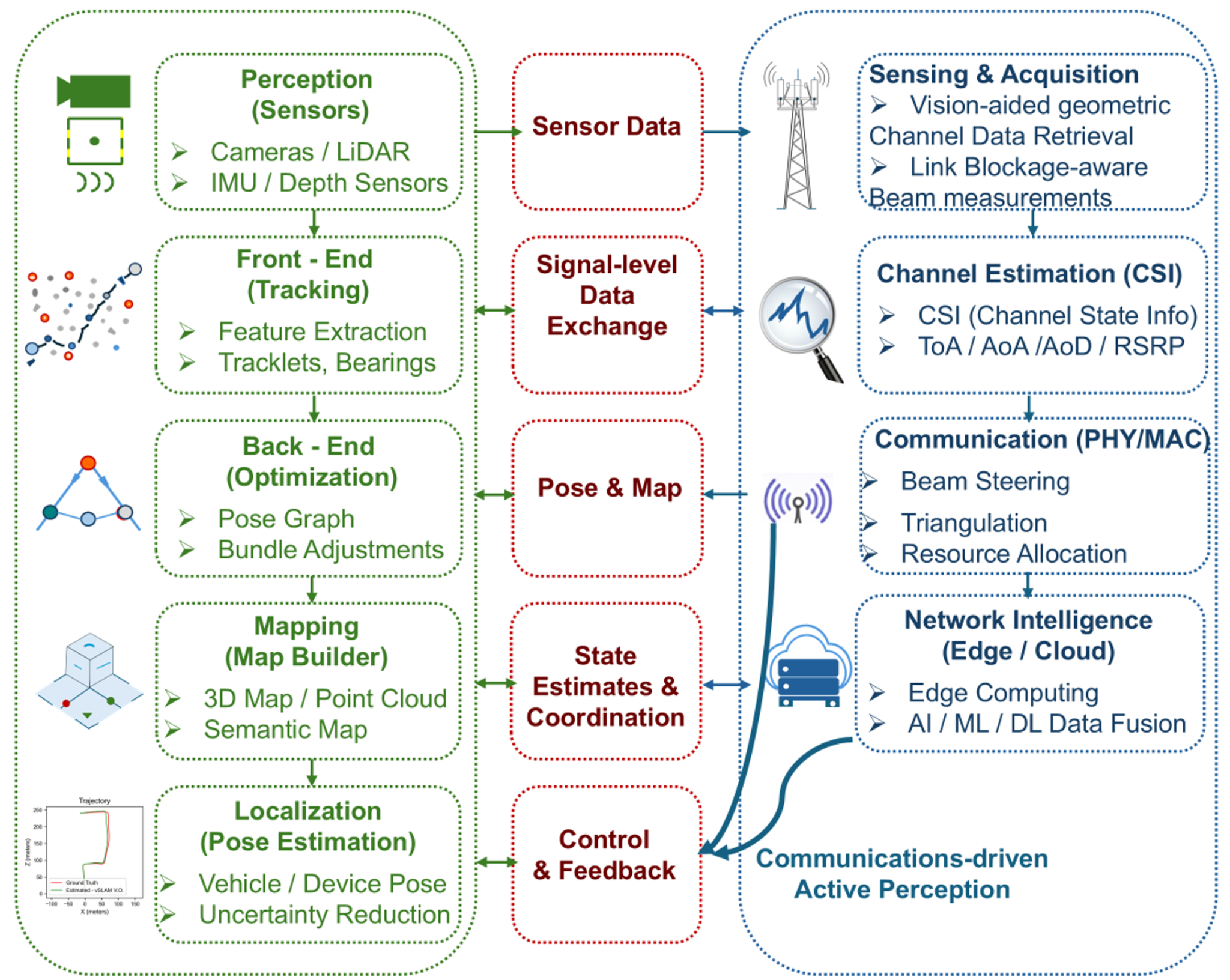}
\caption{{\color{blue}A multilayered, sensor-agnostic structure highlighting the linkages between components of a SLAM pipeline and communications functions where bidirectional relationships are noted as well as the convergence in communications-driven active perception is outlined.}}
\label{fig_2}
\end{figure}

As autonomous systems expand beyond secluded operation, the assumptions that once defined SLAM as a self-contained estimation problem begin to break down. The environments in which these systems operate are increasingly structured by communication, computation, and control processes that interact continuously. In this setting, perception and estimation are no longer bounded by what a single platform can sense but are shaped by the information that flows through its internal and external networks. This shift transforms SLAM from a localized mapping task into a core mechanism that unifies perception and wireless communications with motion control, enabling a system to interpret and coordinate its behavior through shared and adaptive information. {\color{blue} Fig \ref{fig_2} illustrates a multilayered structure highlighting the linkages between components of a SLAM pipeline and related communications functions.} Therefore, this section investigates the underlying principles of a SLAM system and their interactions, linking measurement data acquisition and motion control processes. Specifically, it covers: \textit{Computer Vision Features in SLAM}, which enable the extraction of landmarks and the surrounding scene structure, and provide robust priors that support accurate ego-motion estimation, \textit{Control in SLAM}, being the fundamental mechanism of trajectory tracking, \textit{Passive SLAM}, which derives spatial knowledge from signals and interactions that already permeate the environment, and \textit{Active SLAM}, in which motion and coordination are deliberately guided by the SLAM system itself to improve landmark estimation and achieve the respective computer vision-driven goals such as detection confidence related and/or robotic surveillance-driven or communications-aware perception goals.

A networked autonomous system can be deemed as a mobile, autonomous platform/vehicle, where observation, action, and communication form a unified process of continual feedback that dynamically shapes perception, reasoning, and interaction with the environment. As sensory information flows, state estimates and control signals strengthen each process in turn, forming adaptive cycles that bind perception, motion, and reasoning into a single dynamic loop \cite{Zhang2020}. Through these ongoing exchanges, the circulation of information keeps instantaneous control actions aligned with the system’s cognition, ensuring that every adjustment supports the stability of the overall state. Hence, autonomy arises from the sustained coordination that enables the system to respond coherently to both internal perturbations and external disturbances \cite{Saboia2022}. Formally, it can be defined by the following layers that are also illustrated in Fig. \ref{fig_3}:

\begin{figure}[!t]
\centering
\includegraphics[width=\linewidth]{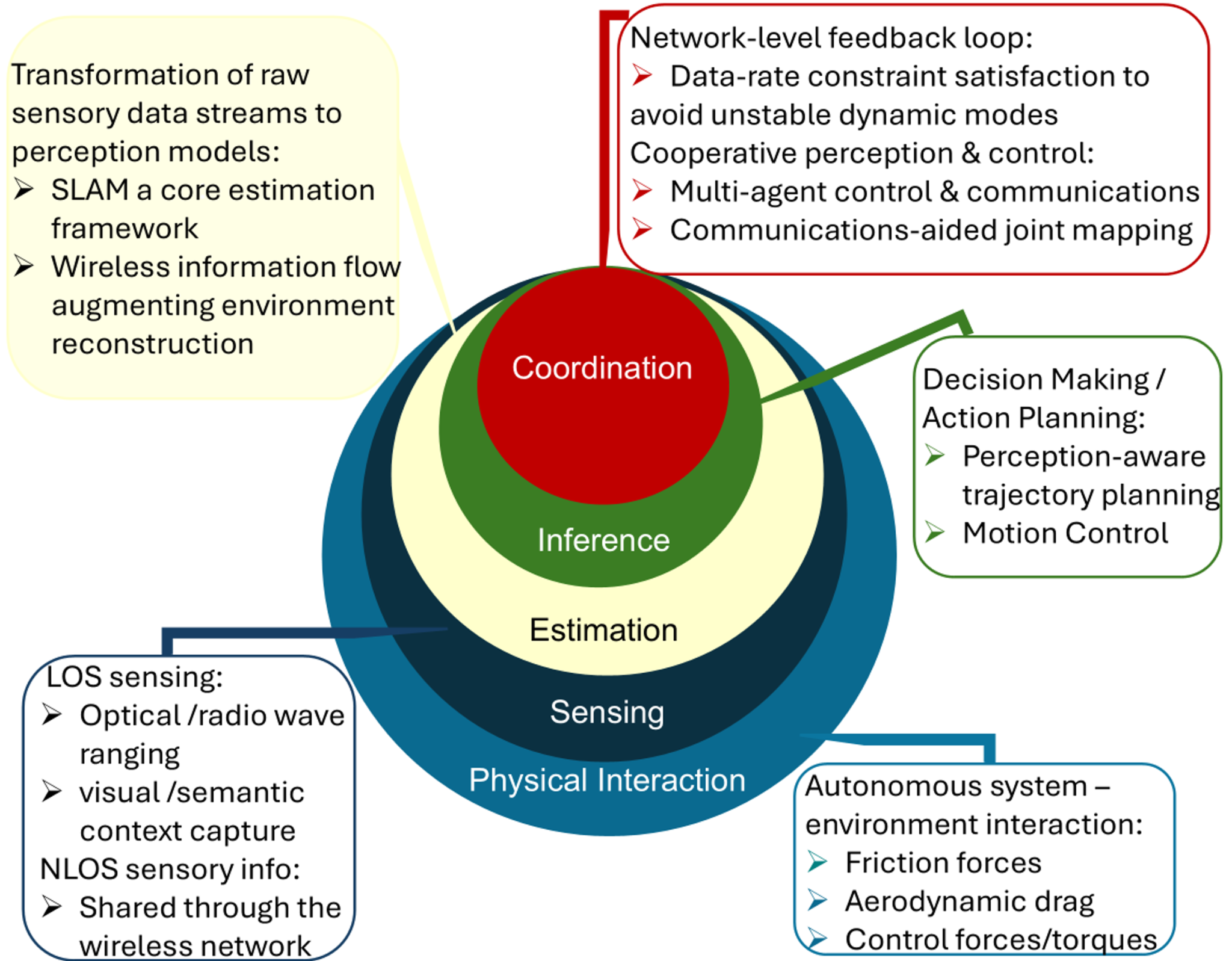}
\caption{{\color{blue}An illustration of the architecture of a networked autonomous system.}}
\label{fig_3}
\end{figure}

\vspace{5pt}
\par\textit{1. Interaction Layer:}
The interaction layer establishes the physical foundation of a networked autonomous system and forms the point where computational decisions take effect in the real world through motion. In this layer, control commands are converted into the forces and torques that generate and regulate movement while compensating for nonlinear behaviors and countering external influences such as surface roughness, longitudinal and lateral friction forces, aerodynamic drag, gravity, and hydrological dynamics. {\color{blue} These functions can be realized through low-level control techniques such as proportional–integral–derivative (PID) control, model predictive control, and nonlinear feedback linearization control, enabling efficient trajectory tracking under time-varying dynamics.} Through these regulating actions, the system maintains stable positioning with its surroundings and achieves the ability to move, manipulate, and interact in a controlled and adaptive manner.

\par\textit{2. Sensing Layer:}
The sensing layer captures how the system and its environment influence one another and transforms these interactions into measurable signals. Through optical, acoustic, radio and ranging modalities, the system observes and reconstructs external structures while {\color{blue}facing} variations in lighting, occlusion, or environmental dynamics. {\color{blue}Key functions may include ranging, extraction of RF/optical features, extraction of semantic context, and determination of acceleration/angular velocity, which are enabled through techniques such as fast Fourier transform (FFT)-based radar signal processing, camera-based feature detection, LiDAR point cloud processing, and inertial sensing.} In connected settings, these sensing pathways also integrate information received through the communication stack, allowing perception to extend beyond direct observation and enhancing awareness through shared sensory data \cite{Urmson2022_CooperativePerception_Survey}. Together, these processes generate a continuous stream of data that describes both the vehicular dynamics and the optical/electromagnetic/mechanical responses of the surrounding structure. Thus, the Sensing Layer provides the perceptual channel through which computational intelligence begins to interpret the environment \cite{Tian2022}.

\par\textit{3. Estimation Layer:}
Based on sensory input, the estimation layer develops a coherent representation of the state of the system and the geometrical structure of its environment. Through the fusion of diverse measurement modalities, it retrieves vehicle kinematics whilst revealing the spatial organization of nearby primitives such as buildings, terrain, and communication infrastructure. {\color{blue}In this layer, state estimation uncertainty quantification and map construction can be approached through probabilistic techniques such as Bayesian filtering (e.g., EKF, particle filter), iterative optimization, and factor graph methods.} SLAM functions as the central estimation framework, transforming data streams into evolving maps and trajectories that preserve consistency up to the retrievable scale over time, while the continual exchange of data through wireless links refines reconstructions and improves accuracy and stability in interconnected conditions \cite{Lajoie2020}. The Estimation Layer thus converts observation into higher-level knowledge, establishing the basis from which reasoning and control emerge.

\par\textit{4. Inference Layer:}
Once a spatial model is formed, the inference layer interprets this knowledge to plan and execute purposeful actions. It connects perception to motion through decision-making, prediction, and high-level control processes that guide the system toward its objectives while adapting to environmental variations. {\color{blue}Its key capabilities can include trajectory planning, risk-aware decision-making, and high-level control, typically achieved through optimization-based motion planning algorithms, sampling-based motion planning algorithms such as the rapidly exploring random tree (RRT), and deep reinforcement learning-based policies.} Planning algorithms and predictive control frameworks within this layer generate trajectories, evaluate outcomes, and continuously adjust execution to maintain safety, efficiency, and stability. By translating spatial understanding into behavior, the Inference Layer closes the perception–action loop and prepares the system for coordinated operation within larger-scale connected contexts \cite{Zhang2020}.

\par\textit{5. Coordination Layer:}
The coordination layer sustains coherence within a networked autonomous system by ensuring that perception, estimation, and control remain aligned through continuous communication across multiple vehicles/robotic platforms \cite{Franceschetti2023}. It governs how information, intent, and spatial knowledge are exchanged so that all subsystems share a consistent understanding of their operational environment. {\color{blue}Key functions may include distributed state estimation and networked control under communication constraints, which can be enabled through techniques such as cooperative perception and localization, and event-triggered communication strategies.} In practice, this coordination relies on low-latency and high-reliability communication frameworks that enable multi-agent estimation and control processes to remain synchronized even under dynamic conditions, as demonstrated in emerging 5G-enabled localization and cooperative perception systems \cite{Urmson2022_CooperativePerception_Survey, Zhou2024_5G_CooperativeAV_Localization}. {\color{blue}From an information-theoretic perspective, the coordination layer is linked to the relationship between communication rate and system stability, where sufficient information flow needs to be maintained to counteract unstable modes in the system dynamics, as formalized by the data-rate theorem in networked control systems \cite{Franceschetti2023}.} Through these mechanisms, local estimates and {\color{blue}control actions} converge to global objectives, supporting joint mapping, adaptive planning, and distributed reasoning \cite{Saboia2022}. The Coordination Layer thus closes the network-level feedback loop, embedding communication as a structural component of autonomy and binding perception, decision, and action into a single coherent process.

\subsection{Computer Vision Features in SLAM}
\subsubsection{Computer Vision-aided SLAM General Outline}
\begin{figure}[!t]
\centering
\includegraphics[width=\linewidth]{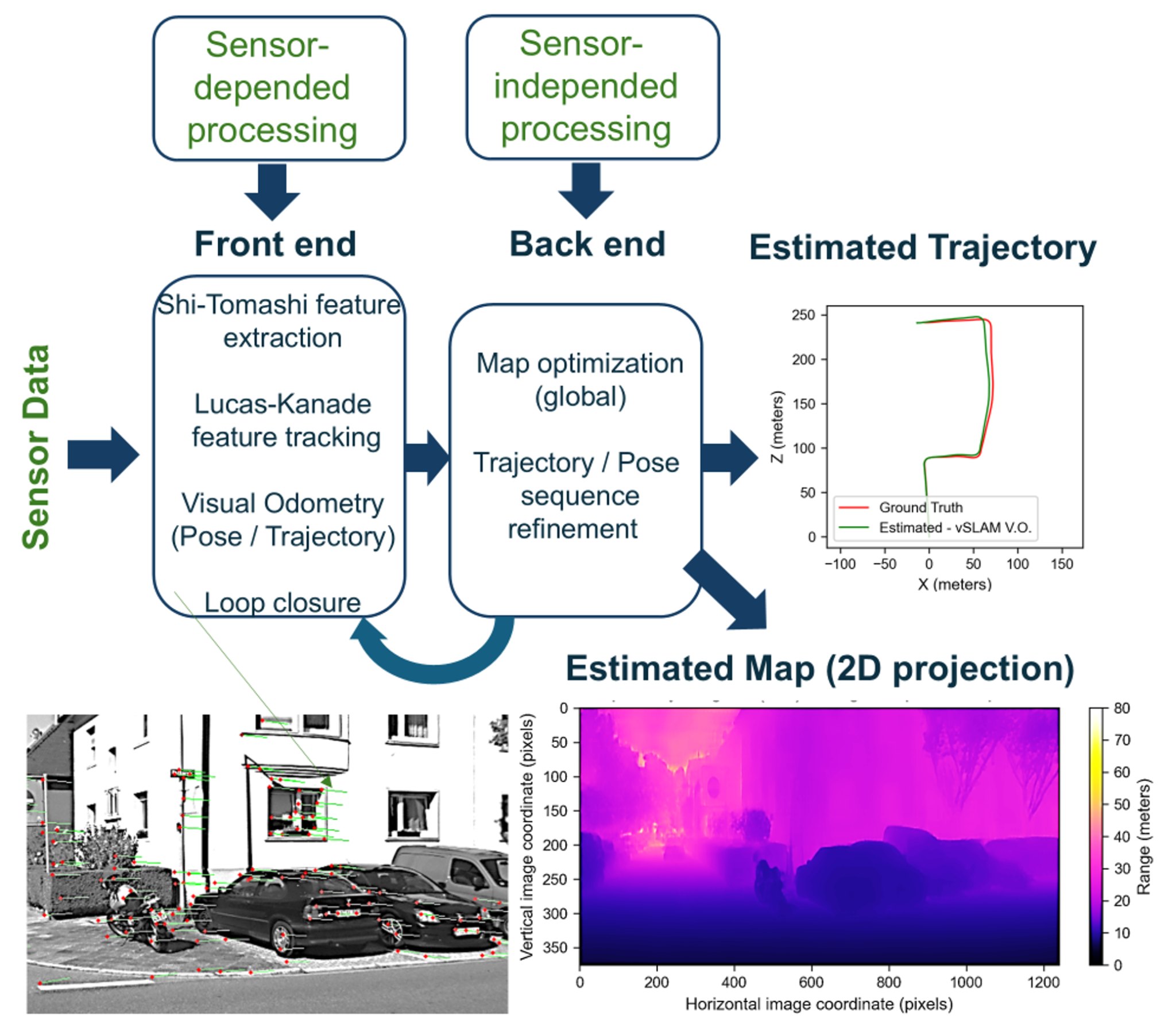}
\caption{{\color{blue}Illustration of the front-end and back-end functionalities in a V-SLAM pipeline. Image frames and Depth Maps are based on the KITTI dataset \cite{Geiger2012}}}
\label{fig_4}
\end{figure}

Classifying SLAM variants based on the nature of the sensing layer data outflow that is built upon, as well as the artificial intelligence primitives applied to them to extract patters, the first class that is prominent due to its great popularity in the Robotics community is V-SLAM. V-SLAM has emerged as a crucial technology for real-time localization and mapping, particularly in environments where external positioning infrastructure is not available. Unlike traditional SLAM methods that rely on LiDAR or inertia measurement units (IMUs), V-SLAM leverages camera-based sensing to extract features from the environment, match them across frames, and estimate motion through geometric and probabilistic models \cite{Zhang2021_VisualSLAM_AutonomousRobot}. Key V-SLAM algorithms such as ORB-SLAM and LSD-SLAM have demonstrated robust performance across various scenarios, yet challenges persist, particularly in dynamic and unstructured environments \cite{Peng2022_DynamicVisualSLAM_MEC_B5G}. V-SLAM algorithms generally consist of four main components: feature extraction, motion estimation, loop closure detection, and map optimization. 
{\color{blue}Fig. \ref{fig_4} shows the components that can be found in a V-SLAM pipeline \cite{Geiger2012}.} Feature extraction is crucial for identifying and tracking distinctive points in the environment, while motion estimation determines the pose of the camera over time. Loop closure techniques improve accuracy by recognizing previously visited locations, reducing accumulated drift, and optimizing the overall map \cite{Peng2022_DynamicVisualSLAM_MEC_B5G}. However, challenges such as occlusion, lighting variations, and real-time computational demands require continuous refinement of these components.

\subsubsection{Fusion of Visual Features with Other Modalities}
Recent research has focused on improving the accuracy of V-SLAM by integrating wireless communication technologies. The fusion of WiFi RSSI signals with ORB-SLAM has been explored to enhance localization stability in indoor environments \cite{Lu2024_ORB_WiFi_Fusion_SLAM}. By leveraging signal strength measurements in conjunction with visual feature extraction, this hybrid approach mitigates feature degradation issues common in texture-less areas. Additionally, 6G networks have been proposed as a means to enhance V-SLAM processing, enabling real-time location recognition and cloud-based SLAM computations for large-scale environments \cite{Zhang2024_SLAM_6G_Localization}. Advancements in 5G and beyond wireless networks have further contributed to the development of network-assisted SLAM, where mobile devices offload computationally intensive tasks to edge servers. This offloading reduces latency, conserves battery life, and allows for collaborative SLAM frameworks, where multiple devices share mapping data in real time \cite{Zhang2024_SLAM_6G_Localization}. Such an approach is particularly useful for indoor environments, where occlusions and multipath effects often limit WiFi-based localization techniques \cite{Segura2011_UWB_Localization_SLAM}.

The integration of depth cameras has been a major advancement in V-SLAM, particularly for robotics applications. Depth-enhanced SLAM allows more accurate reconstructions, making it suitable for mobile robot navigation \cite{Yin2023_Localization_DepthCamera}. Additionally, monocular vision-based SLAM for UAV localization has been explored to provide GPS-independent navigation solutions for drones in GPS-denied environments \cite{Nie2022_IndoorLocalization_UAV_Vision}. Another key area of research involves edge computing and particularly Multi Access Edge Computing (MEC) architectures for SLAM, where real-time processing is distributed across multiple computing nodes to reduce latency and improve computational efficiency \cite{Peng2022_DynamicVisualSLAM_MEC_B5G}. Such techniques are especially useful for mobile robots in smart cities, where large-scale dynamic environments pose significant computational challenges \cite{Kuang2022_MonocularVisualInertialSLAM_6G}. The fusion of vision with inertial sensors (VIO-SLAM) is another method that enhances robustness, especially in environments where feature tracking is difficult. By combining IMU readings with camera data, these approaches improve motion estimation and mitigate tracking failures in low-texture or high-speed motion scenarios \cite{Zhang2021_VisualSLAM_AutonomousRobot}.

In augmented reality (AR), multi-view data capture techniques along with learned point features have been used to reconstruct dynamic objects in real-time \cite{Bortolon2021_MultiView_Reconstruction_AR, Kirillov_2020_CVPR}. Such developments have profound implications for interactive applications, including AR-assisted SLAM for navigation and remote collaboration. The application of SLAM-based AR in industrial settings is also gaining traction, enabling workers to interact with overlaid digital content for training and maintenance purposes \cite{Wang2022_AR_IndustrialTraining_Maintenance}. Additionally, research has explored multi-user SLAM environments, where multiple AR users share a common spatial understanding through real-time collaborative mapping. This approach enables seamless interaction between human operators and autonomous systems, improving efficiency in robotic teleoperation, remote assistance, and mixed-reality applications \cite{Rodic2015_ServiceRobot_EmbeddedPersonality}. 

\subsubsection{Computer Vision-aided SLAM Challenges}
Despite the discussed advantages that V-SLAM offers, several notable challenges remain:

\begin{list}{}{}
\item{a. Scalability - Handling large-scale environments while maintaining map consistency.}
\item{b. Real-Time Processing - Reducing computational overhead for embedded and mobile platforms.}
\item{c. Hybrid SLAM Systems - Exploring more efficient fusions of visual, RF, and IMU data to improve localization performance in diverse environments.}
\end{list}

Emerging trends in V-SLAM research include the development of AI-enhanced feature extraction, multimodal SLAM architectures, and real-time cloud-based SLAM frameworks. The integration of 6G-enabled SLAM is expected to play a crucial role in future advancements, providing ultra-low-latency data exchange and enhancing the accuracy of collaborative SLAM applications \cite{Zhang2024_SLAM_6G_Localization}. Additionally, research in bio-inspired SLAM models is gaining interest, leveraging neuromorphic computing and spiking neural networks to mimic human vision and perception mechanisms. These approaches aim to achieve more power-efficient and adaptive SLAM solutions, especially for low-power robotics and edge devices \cite{Rodic2015_ServiceRobot_EmbeddedPersonality}. By continuing to refine sensor fusion, DL-based SLAM, and large-scale mapping solutions, the field of V-SLAM is poised to enable new frontiers in autonomous navigation, assistive technology, and intelligent augmented environments.

\subsection{Control in SLAM}
\subsubsection{Kinematic and Dynamic Models Outline}
\begin{figure}[!t]
\centering
\includegraphics[width=\linewidth]{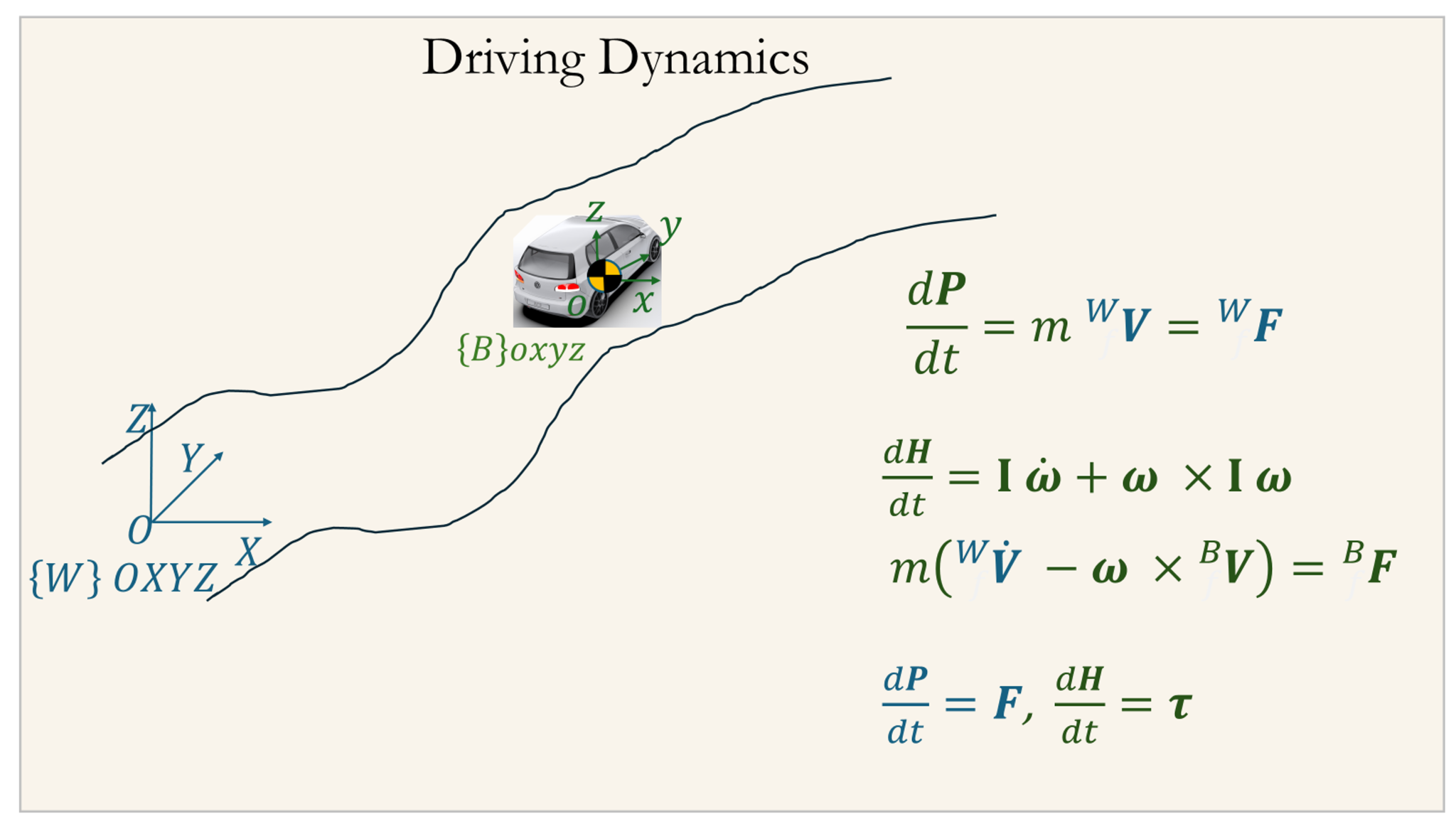}
\caption{The dynamics of an autonomous car using Newton-Euler equations and the transport theorem that relates the body frame quantities to earth-fixed frame ones.}
\label{fig_5}
\end{figure}

Motion planning and control are essential functionalities since a robot is not deemed fully integral without the ability to reason about and then physically interact with its environment. There is a large emphasis on developing multi-sensor-based, perception–aware planning and control pipelines in applications such as automated manufacturing, transportation, farming, delivery of goods, medical robotics, inspection, surveillance, and extremely precise mapping and prediction for emergency management. Research in robotic systems such as autonomous vehicles \cite{Sahoo2019_AUV_Advancements} is marked by increased popularity because of their extensive applications in fields from military to science. The kinematic model of a mobile robot is the mathematical relationship between the inertial, non-inertial frame and the links of a robot which defines the position, velocity and acceleration of different parts of the robot with respect to some frame of reference. The dynamic model refers to the nonlinear differential equation that relates inertial, Coriolis, centripetal, gravitational and damping, i.e., friction or viscous damping forces and moments with the derivatives of the position and velocity of the robot. This mobile robot mathematical formulation can be easily extended to multiple rigid body systems. Typically, a simple mobile robot is often described by a kinematic model that correlates the body-fixed frame and the earth-fixed frame as shown in Fig. \ref{fig_5}. The body-fixed frame is attached to the geometrical center of the vehicle with axes in the directions of surge, sway and heave, respectively. Earth-fixed or inertial frame coincides with the north-east-down directions and fixed to a point on the water surface.

\subsubsection{Low-Level Control and Estimation}
Concerning the low-level control of autonomous vehicles, classic control techniques such as PID control are commonly used. Apart from simple controllers, researchers have developed more sophisticated laws such as nonlinear control, adaptive control, sliding mode control, and neural network control to regulate a multitude of the vehicle's degrees of freedom. Fuzzy logic controllers are popular control techniques. In \cite{Vo2010TurningFishRobot}, the sliding mode controller (SMC) and the fuzzy sliding mode controller (FSMC) were proposed to achieve the straight and turning motion of the robot. The feasibility and quality of the controllers were verified using numerical simulations. Aerial robots, when coupled with an IMU, can be controlled in an underactuated manner to perform agile flying maneuvers. The original quadrotor literature mostly focused on the modeling, state estimation, and low-level control of the vehicle while assuming that the localization, position, and orientation estimation required for feedback control is externally and readily available (e.g., from precise GNSS receivers or indoor motion capture systems). In fact, state estimators were eventually developed to close the control feedback loop without the need for external positioning-based devices, but this development occurred well after the control pipeline had been established \cite{Cristofalo2020_VisionBasedControl_3DReconstruction}. The co-design occurs where the control influences the belief prediction step of a Gaussian State Estimator such as an EKF, often being central to the system that addresses SLAM as previously discussed.

From the perception-aware control point of view, a critical challenge lies in the fact that the control vector must be computed, given any inaccuracies left over from the perception process. For example, pose graph optimization methods are required to compute the best possible estimate of the robot poses taking into consideration potentially corrupted relative pose measurements. Supplementary to the perception related challenges, the control computation must also be feasible for a robot’s onboard computer which can be challenging when dealing with massive state spaces or complex systems. In SLAM, the size of the vector representing the world grows with the addition of each (RF-based, mmWave-based, visual) measurement of a salient world feature. Planning motion timing laws, geometric paths, or control inputs in these vastly growing indoor or outdoor environments can become unmanageable if not addressed.

\subsubsection{Automated and Cooperative Driving}
Connected autonomous cars, either up to a specific level of automation provided by advanced driver assistance systems (ADAS) or to the level of complete autonomy (SAE Level 5), have certainly attracted the interest of the society worldwide as well as industry and academia. The latter might be explained by the fact that road accidents account for a considerable loss of lives worldwide. From the perspective of integrated sensing, control, and communications (ISCAC), existing autonomous vehicle localization methods typically rely on the cooperation between onboard sensors, cellular infrastructure elements, and vehicular networks to satisfy the demanding localization requirements. Advances in 5G networks can revolutionize cellular localization performance while providing the capacity to handle massive sensor data and enabling reliable vehicle-to-everything (V2X) information flow at high data rates with minimal latency \cite{Zhou2024_5G_CooperativeLocalization_Survey}. Therefore, the vehicular network deployment is of great importance, so that stable performance is acquired to compensate for uncertainties related to neighboring vehicles, environmental and road conditions. Moreover, the capability of seamless transmission of huge amount of data at an optimal rate is of paramount importance \cite{Zhou2024_5G_CooperativeAV_Localization}.

Considering that the significant amount of data collected by autonomous perception sensors can also lead to significant power consumption in the vehicle, in \cite{Katare2023_ApproximateEdgeAI_AutonomousDriving}, the perspective of energy efficiency in autonomous driving is surveyed. To reduce energy consumption in electric vehicles so that a barrier is established for safety-critical maneuvers that require instantaneous acceleration/braking/yaw control, the authors highlight AI model compression techniques for the vehicle perception systems. Moreover, the model approximation techniques are reviewed, namely: (i) quantization of vehicular sensory information, (ii) sparsification in the sense of removing sensory data vector elements or approximating them in a lower-dimensional space, e.g., via low-rank-approximation techniques, (iii) knowledge distillation to approximately represent a larger DNN model in compressed/reduced form.

Networked control systems in autonomous driving are characterized by attributes such as the deployment of protocols that standardize vehicle communications to ensure V2V connection and the optimization of the roadside unit configuration to avoid the inefficient distribution of 5G BSs \cite{Zhou2024_5G_CooperativeAV_Localization}. Vehicular network designs are expected to efficiently allocate 5G resources in order to avoid conflict between localization and communication applications, including transmission power, bandwidth, and rate \cite{Zhou2024_5G_CooperativeAV_Localization}. An effective approach is to establish an impartial packet priority system that considers both latency and reliability requirements, followed by the strategic allocation of additional resources to critical applications. Autonomous vehicle control networks face significant risks, including unauthorized access, data breaches, and privacy violations. These security and privacy vulnerabilities could lead to potential traffic accidents, highlighting the urgent need to address them for a robust 5G-V2X system. Ensuring the integrity and confidentiality of all transmitted data is crucial to enhancing the overall security and privacy of the 5G-V2X localization framework. Implementing countermeasures such as message authentication and cryptographic key management can further safeguard shared information.

\subsection{Passive SLAM for Networked Autonomous Systems}

\textit{Passive SLAM} denotes the intrinsic ability of a networked autonomous system to infer motion and spatial structure from the same data streams that sustain its operation, using vision data, wireless signals, and shared measurements without deliberate exploration or additional probing transmissions \cite{Zhang2024_SLAM_6G_Localization}. System level-wide, this capability is realized through the combined operation of the sensing, state observation, and high-level control and coordination modules, where sensing captures environmental observations through visual and radio modalities, estimation fuses these into probabilistic spatial models that evolve with each update \cite{Peng2022_DynamicVisualSLAM_MEC_B5G, Kuang2022_MonocularVisualInertialSLAM_6G}, and the high-level control and coordination modules plan the path and motion trajectories whilst exchanging state information among distributed inter-vehicular or intra-vehicular modules \cite{Cao2024}. Consequently, these coupled processes allow the networked autonomous system to maintain an accurate state estimate that remains stable under suitable modeling conditions even without explicit exploratory actions. Therefore, passive SLAM represents a fundamental attribute of networked autonomy, showing how continuous estimation and communication preserve collective spatial awareness throughout the system \cite{Saboia2022}.

During measurement data {\color{blue}acquisition} step of an autonomous system, wireless quantities such as time of arrival, angle of arrival, carrier phase, and multi-path structure, together with computer vision-related quantities such as range, depth, and pose are {\color{blue}retrieved}. Because these quantities characterize the relative geometry between the {\color{blue}agents} and their environment, they can be directly incorporated into the estimation process without additional sensing effort. Hence, communication feedback signals that preserve connectivity also convey geometric information that supports spatial inference. {\color{blue}An interesting capability they may encompass is providing the source of metric information that aids scale recovery in the context of monocular V-SLAM}. {\color{blue}In a general setting, their dual role of carrying information and embedding geometry} is analogous to the estimation based on essential matrix decomposition in computer vision, where {\color{blue}3-D landmarks}, inferred from multi-view feature correspondences related via the {\color{blue}epipolar geometry}, can be used to determine camera translational and angular velocities\cite{Scaramuzza2011}. Furthermore, the wireless propagation channel participates in estimation, since variations in signal properties reveal environmental geometry while reflecting network conditions such as latency and link reliability. Experimental and analytical studies in multipath-based SLAM confirm that reflections and virtual anchors contribute to joint trajectory and feature estimation, demonstrating that the propagation channel retains persistent geometric information about the scene \cite{Leitinger2023_DataFusionMPSLAM}. In addition, research in decentralized SLAM shows that local maps can remain consistent when only selected loop closures or reduced factor-graph data are exchanged, with global alignment achieved through distributed optimization rather than centralized fusion \cite{Lajoie2023_SwarmSLAM}. Finally, developments in ISAC reveal that modern wireless systems are being designed to expose localization and channel-state information as intrinsic capabilities, embedding spatial inference directly within the operation of the network \cite{Liu2022_ISAC_DualFunctional, Zhang2025_IntelISAC}.


Attributing the aspects of perception-aware communication and communication-efficient mapping and localization methods, it has been demonstrated that spatial consistency can be preserved by transmitting only selected keyframes, compressed features, or reduced factor-graph structures, since each exchanged message is evaluated according to its information contribution rather than sent at a fixed rate \cite{Ge2025_BandwidthAwareMapDistillation}. Furthermore, adaptive estimation strategies show that update frequency, map resolution, and feature density can be adjusted in response to the instantaneous {\color{blue}measurements} of the wireless infrastructure and the available energy budget, allowing the networked autonomous system to maintain estimation stability even under fluctuating link quality \cite{KiranJot2021}. 

{\color{blue}To provide a unified mathematical framework where pose estimation, motion input, and geometric channel-related quantities are integrated into a single estimation model, the problem of V-SLAM for a vehicle equipped with a monocular camera operating in an urban environment is considered. Let $I_{k-1}$ and $I_k$ denote two consecutive image frames acquired at discrete time instants $k-1$ and $k$.

\subsubsection{Feature Detection and Tracking}

At initialization ($k=0$), the pixel coordinates of salient features in the image signal that can be extracted using a geometric primitive detection method, denoted (per-feature) by

\begin{equation}
\mathbf{x}_{i,0} =
\begin{bmatrix}
x_{i,0},\
y_{i,0}
\end{bmatrix}^\top.
\end{equation}

Feature detection may occur at the beginning of the autonomous vehicle navigation process. The future pixel location of each feature in the subsequent image can be predicted using a feature tracking algorithm. The Lucas-Kanade (LK) optical flow is a tracking algorithm known for its efficiency realizing the pyramid sampling mechanism for coarse-to-fine estimation as well as its real-time performance \cite{LucasKanade1981, Bouguet2001PyramidalLK}. According to this algorithm, for each feature $i$, its displacement $\mathbf{d}_{i,k}$ is obtained by minimizing:

\begin{equation}
\mathbf{d}_{i,k} =
\arg\min_{\mathbf{d}}
\sum_{\mathbf{x} \in M_i}
\left( I_k(\mathbf{x} + \mathbf{d}) - I_{k-1}(\mathbf{x}) \right)^2,
\end{equation}
where $M_i$ denotes a local image patch. It is remarked that constant brightness is assumed between two consecutive frames. The tracked feature location is then

\begin{equation}
\mathbf{x}_{i,k} = \mathbf{x}_{i,k-1} + \mathbf{d}_{i,k}.
\end{equation}

The initialization of the features to track with the LK algorithm can also occur at preselected fixed time intervals or upon loss of the tracks. 

\subsubsection{Epipolar Geometry and Relative Pose Estimation}

Feature points corresponding to the same 3-D structure, as tracked from two consecutive camera viewpoints are constrained each to be in the epipolar plane formed by the two locations of the camera center and the respective 3-D point. The latter intersects each image plane at the two measurement acquisition instances. Solving the epipolar constraint can lead to the retrieval of the pose of the autonomous vehicle, under certain assumptions. 

To derive the epipolar constraint model, the image coordinates of a tracked feature at time $k$ are normalized by the horizontal and vertical focal lengths $f_x$ and $f_y$, respectively, yielding:
\begin{equation}
\tilde{\mathbf{x}}_{i,k} =
\begin{bmatrix}
(x_{i,k} - c_x) / f_x,\
(y_{i,k} - c_y) / f_y,\
1
\end{bmatrix}^\top.
\end{equation}

$c_x$ and $c_y$ correspond to the pixel coordinates of the principal point, often taken as the center of the image.

The essential matrix $\mathbf{E}_k$ satisfies the epipolar constraint:
\begin{equation}
\tilde{\mathbf{x}}_{i,k}^\top \mathbf{E}_k \tilde{\mathbf{x}}_{i,k-1} = 0.
\end{equation}

Using robust parameter estimation methods and singular value decomposition (SVD), $\mathbf{E}_k$ is decomposed as:
\begin{equation}
\mathbf{E}_k = [\hat{\mathbf{t}}_k]_\times \hat{\mathbf{R}}_k,
\end{equation}
yielding the relative rotation in the special orthogonal group $(3)$ $\hat{\mathbf{R}}_k \in SO(3)$ and the translation direction $\hat{\mathbf{t}}_k \in \mathbb{R}^3$, defined up to an unknown scale. For brevity, $\hat{\mathbf{R}}_k$ and $\hat{\mathbf{t}}_k$ denote the relative rotation and translation from the camera coordinate frame at $k-1$ to the frame at $k$.

The corresponding homogeneous transformation is:
\begin{equation}
\hat{\mathbf{H}}_{k} =
\begin{bmatrix}
\hat{\mathbf{R}}_k & \hat{\mathbf{t}}_k \\
\mathbf{0}^T & 1
\end{bmatrix}.
\end{equation}

Pose estimates w.r.t the initial camera coordinate system, considered the inertial frame hereinafter, are obtained recursively:
\begin{equation} 
\mathbf{P}_k = \mathbf{P}_{k-1} \hat{\mathbf{H}}_{k}^{-1}.
\end{equation}

\subsubsection{Landmark Reconstruction}

Given the feature correspondences across camera views, the apparent motion of a 3-D landmark $\mathbf{X}_i$, as viewed from a moving camera, is:

\begin{equation}
\mathbf{\dot{X}}_{i}(t) = \mathbf{\Omega}(t) \mathbf{X}_{i}(t) + \mathbf{v}(t).
\end{equation}

The discrete-time form is:

\begin{equation}
\mathbf{X}_{i, k} = e^{\mathbf{\Omega}_{k} dt} \mathbf{X}_{i, k-1} + \mathbf{v}_{k} dt,
\end{equation}

where $\mathbf{\Omega}_{k} = [\boldsymbol{\omega}_{k}]_\times$ is the skew-symmetric matrix in $SO(3)$ that describes the angular velocity the camera exerted, responsible for the change of orientation from the coordinate frame at $k-1$ to the coordinate frame at $k$, and expressed in the former coordinate frame. $\mathbf{v}_{k}$ represents the translational velocity of the camera from the coordinate frame at $k-1$ to the coordinate frame at $k$, expressed in the former coordinate frame.

Using the rotation and translation components of the pose estimate and exploiting the fact that $\hat{\mathbf{R}}_{k} \approx e^{\mathbf{\Omega}_{k} dt}$ and $\hat{\mathbf{t}}_{k} \approx \frac{1}{\lambda_k} \mathbf{v}_{k} dt$:

\begin{equation}
\hat{\mathbf{X}}_{i, k} = \hat{\mathbf{R}}_{k} \hat{\mathbf{X}}_{i, k-1} + \hat{\mathbf{t}}_{k},
\end{equation}

where $\hat{\mathbf{X}}_{i, k} = \frac{1}{\lambda_k} \mathbf{X}_{i, k}$ represents the location of each tracked landmark up to an arbitrary scale. Scale recovery followed by the determination of the landmarks' locations w.r.t an inertial coordinate frame results in a metric raw map.

\subsubsection{Scale Ambiguity and Metric Motion Input}

The essential matrix decomposition provides the relative translation up to scale:
\begin{equation}
\mathbf{t}_k = \lambda_k \hat{\mathbf{t}}_k,
\end{equation}
where $\lambda_k > 0$ is an unknown factor.

To resolve this ambiguity, one can acquire via the odometry sensors the following metric displacement that, in the dynamics of both the autonomous vehicle and the landmarks as seen from the moving camera, serves the role of the motion control input \cite{thrun2005probabilistic}:
\begin{equation}
\Delta \mathbf{p}_k \in \mathbb{R}^3,
\end{equation}
providing the true translational motion between frames in metric units. $\lambda_k$ is estimated by solving:
\begin{equation}
\lambda_k =
\arg\min_{\lambda > 0}
\left\|
\Delta \mathbf{p}_k - \lambda \hat{\mathbf{t}}_k
\right\|_2^2.
\end{equation}

Once $\lambda_k$ is obtained, the pose and the landmarks can be expressed in metric units.

\subsubsection{Sources of Metric Motion Input}

The metric displacement $\Delta \mathbf{p}_k$ can be obtained from different sensing modalities:

\paragraph{Inertial Odometry}

Inertial measurements provide acceleration $\mathbf{a}_k$, from which velocity and displacement can be estimated via double integration, numerically approximated as:

\begin{equation}
\Delta \mathbf{p}_k \approx \mathbf{v}_{k} \Delta t.
\end{equation}
\begin{equation}
\mathbf{v}_k = \mathbf{v}_{k-1} + \mathbf{a}_{k} \Delta t,
\end{equation}
This procedure is often prone to drift, raising the need for sensor fusion (e.g. through Kalman filtering) along with bias compensation and gravity removal.

\paragraph{Wireless Communications-based Odometry}

Metric position estimates can alternatively be obtained from communication signals exchanged with multiple base stations (gNBs) of known positions $\mathbf{s}_b$. Measurements such as time-of-arrival (ToA), round-trip time (RTT), angle-of-arrival (AoA), and received signal strength (RSS) can satisfy the following measurement model \cite{Italiano2025_5G_Positioning_Tutorial}:

\begin{equation}
z_{b,k} = h_b(\mathbf{p}_k, \mathbf{s}_b) + n_{b,k},
\end{equation}
where $\mathbf{p}_k$ denotes the position of the autonomous vehicle and $n_{b,k}$ measurement noise.

By combining measurements from multiple gNBs, a position estimate $\mathbf{p}_k^{\mathrm{RF}}$ can be obtained via multi-lateration/angulation \cite{Italiano2025_5G_Positioning_Tutorial}. The corresponding metric displacement is:
\begin{equation}
\Delta \mathbf{p}_k^{\mathrm{RF}} =
\mathbf{p}_k^{\mathrm{RF}} - \mathbf{p}_{k-1}^{\mathrm{RF}}.
\end{equation}

This quantity could be used directly in the scale estimation problem:
\begin{equation}
\lambda_k =
\arg\min_{\lambda > 0}
\left\|
\Delta \mathbf{p}_k^{\mathrm{RF}} - \lambda \hat{\mathbf{t}}_k
\right\|_2^2.
\end{equation}
}

\begin{figure}[!t]
\centering
\includegraphics[width=\linewidth]{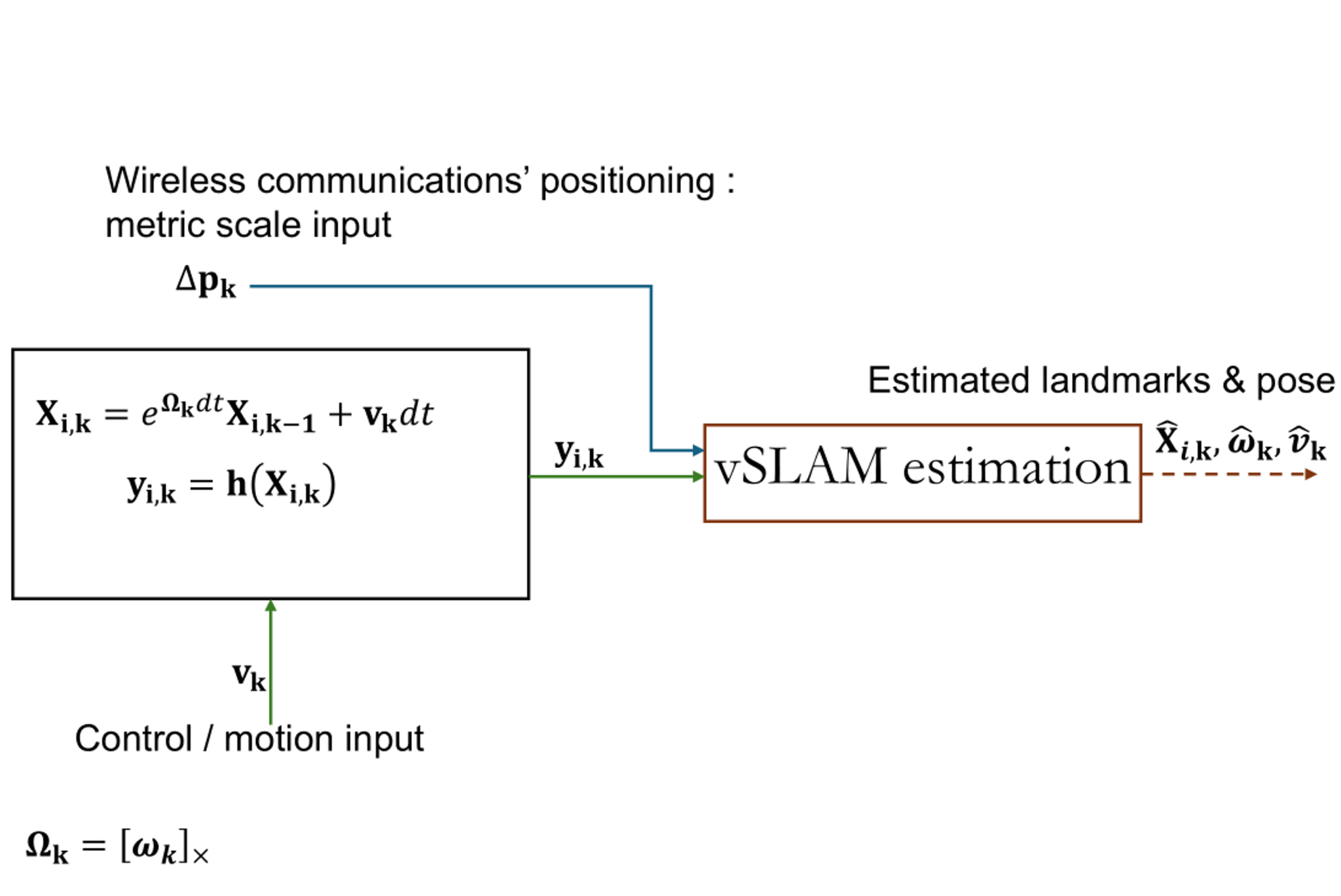}
\caption{{\color{blue}Abstract representation of V-SLAM as a system subject to the apparent motion dynamics $\mathbf{\dot{X}}_{i}(t)$, fusing image features with motion input from inertial and/or RF-based odometry, and providing landmark and pose estimates.}}
\label{fig_6}
\end{figure}

{\color{blue}Built upon algorithmic pipelines with the same core elements as in the framework described above and also illustrated in Fig. \ref{fig_6}, the perception module of a networked autonomous system {\color{blue}may incorporate and augment} the vision-based inference mechanisms that have shaped robotic SLAM. Networked implementations reinterpret them within a distributed framework, where visual cues are fused with motion inputs, radio-based signals, and network state information to maintain situational awareness under dynamic conditions.} 

It is noted that the perception capabilities of a networked autonomous system can be extended beyond its physical components in the context of virtual reality and digital twin modeling that can synchronously evolve with the real system. Within this expanded loop, spatial models generated by sensing and estimation are mirrored in a virtual domain that remains aligned with the physical world through the same wireless channels that coordinate interaction among networked agents. This correspondence allows the system to refine its map and state estimate even when direct observations are unavailable. In practical implementations, live radio and mapping data are streamed to edge or cloud processes that reconstruct the surrounding space, creating a continuously updated digital twin of the operating network \cite{RodaSanchez2025_NetworkDigitalTwin_5G}. In parallel, physics-informed learning frameworks developed within simulated communication environments demonstrate that networked autonomous systems trained on channel metrics such as delay spread, path loss, and angle of arrival can operate directly in real deployments without retraining, revealing that domain adaptation arises and the sim-to-real gap closes naturally when learning over physically consistent propagation statistics \cite{Li2025_PhysicsInformedRL_mmWave}. Additionally, virtual testbeds of radio propagation, object detection and tracking pipeline testing, and signal interaction allow visualization and analysis of how network dynamics influence estimation performance, providing an interpretable layer for examining the perception-wireless communications nexus\cite{Nisiotis2024_WirelessPropagationDT}. Therefore, Passive SLAM in networked autonomous systems evolves toward a hybrid regime in which estimation unfolds simultaneously across physical and virtual domains, allowing resilient estimations to emerge as intrinsic outcomes of communication-driven and domain-driven perception unified within a learning model that supports autonomy.

\subsection{Active SLAM for Networked Autonomous Systems}

The concatenation of networked autonomous systems-abundant, visual and RF data used for agent pose as well as landmark position estimation was discussed, without any assumptions on the bidirectional relationship between any applied control input and the output observation. The problem of controlling the agent’s motion to minimize the uncertainty of its representation and localization on the map is termed active SLAM \cite{cadena2016slamfuture}. In her study, Bajcy \cite{Bajcsy1988_ActivePerception} first stated that an observer is considered active when engaged in any activity whose purpose is to control the geometric parameters of the sensory system in order to achieve a sensing/perception objective. In the same paper, it is stated that an actively observing robot is characterized by its ability to determine the “why” of its behavior and by its attempts to answer at least one of the “what”, “how”, “where” and “when” questions throughout the control of each behavior it exhibits. In 1988, Bajcsy in her paper defined active perception as the study of modeling and control strategies for perception. The modeling strategies are both local model-based and global model-based. The former focus on aspects of the sensor model, such as lens distortion, spatial resolution, and bandpass filtering of received signals. The latter focus on the overall system outcome and predict how individual modules interact.

A popular framework for active SLAM is focused around best future action selection among a finite set of alternatives. This family of active SLAM algorithms flows in three main steps \cite{Blanco2008_UncertaintySLAM_RBPF}: a) The agent highlights possible locations to explore or exploit the so called vantage locations in its current map approximation, b) The agent computes an utility metric for each paid vantage point visit and picks the action with the highest utility and c) The agent carries out the selected action and decides if it is deemed necessary to continue or to terminate the task.

Control-theoretic approaches have been focused on the integration of an optimization-based framework into the active estimation problem. A promising choice for optimization-based control is the model predictive controller. Realizing a sophisticated control algorithm on a networked autonomous system that is bounded by hard real-time constraints whilst moving in an agile manner at a high speed, such as a UAV, is potentially challenging. The latter arises from the fact that changes occur when new features or obstacles are detected and when the estimates in a state observer such as the EKF are updated. Model predictive control frameworks are well suited to such systems with highly dynamic environments. Due to the fact that the underlying algorithms are typically simple and fast, continuous re-planning is enabled in order to incorporate feedback and up to date knowledge of the system state and environment \cite{Leung2006_ActiveSLAM_MPCAttractor}. The working principle of model predictive control suggests: a) predicting the following motion states of the robot over a time horizon, b) computing the control input to achieve a desired trajectory over the next horizon time steps whilst regulating error states caused by external disturbances to zero throughout the optimal solution to a non-convex problem (in the general case) or its locally (over a defined state and input region) quadratic approximation, and c) applying only the next step computed control input to the robot and when new sensory data or estimates of them become available, repeating the optimization step.

As SLAM becomes more tightly coupled with sophisticated nonlinear control laws, it can be easily deduced that an upcoming question may concern the particular metric with which the control actions need to be optimized. Two prominent design decisions involve: a) maximizing the information gain, considering motion state evolution in a way that is analogous to node creation through splitting in decision trees and balancing entropy reduction with exploratory efficiency and b) minimizing estimation uncertainty, achieved through mathematical concepts derived from Gaussian or Kalman-filtering formulations, such as minimizing the trace of the posterior covariance matrix or maximizing the smallest eigenvalue of the observability Gramian matrix  \cite{Feder1999_AdaptiveMobileRobotNavigationMapping} addressed the active-SLAM problem by applying a greedy strategy: at each step the robot was instructed to select the action that maximizes the expected information gain based on to-date knowledge. Notably, they observed that a longer planning horizon could yield further performance gains. Cristofalo \cite{Cristofalo2020_VisionBasedControl_3DReconstruction} developed UAV control algorithms to recursively estimate the 3-D location of an object point expressed in the onboard camera frame. The extended and unscented Kalman filters (UKF) were implemented, incorporating the state space model of the motion of a point in a moving frame and the perspective projection model as the state transition and measurement models, respectively. The UAV velocity vector in the frame of the monocular camera was chosen as the control input. To improve the way a 3-D point of interest was perceived, the process model equations were solved with respect to the sought control vector that minimizes the trace of the posterior covariance matrix, in an online gradient-based optimization scheme, inducing UAV motion commands along the direction of the steepest descent.

{\color{blue}

\subsubsection{Communications-driven Active Perception}

To provide a unified mathematical framework where a wireless communication objective is viewed as an uncertainty reduction objective and tackled through selecting motion inputs that improve landmark estimation accuracy, the problem of active V-SLAM for a vehicle equipped with a monocular camera operating in an urban environment is considered. 

Let $\mathbf{X}_{i,k}$ denote the coordinates of the $i$-th tracked 3-D landmark expressed in the camera coordinate frame at time $k$. In the communication-aware active perception case, the landmark may correspond to a tracked user equipment (UE) or transmitter whose position estimate is important for optimal beam alignment. The apparent landmark dynamics induced by camera motion 


\begin{equation}
\mathbf{X}_{i,k} \approx \left(\mathbf{I}_{3}+\mathbf{\Omega}_{k}dt\right)\mathbf{X}_{i, k-1}
+ \mathbf{v}_{k} dt,
\end{equation}

can be written as:

\begin{equation}
\mathbf{X}_{i, k}
=
\mathbf{f}(\mathbf{X}_{i,k-1},\mathbf{u}_k)
+
\mathbf{w}_{i,k},
\end{equation}

where $\mathbf{u}_k$ denotes the velocity/control input $\mathbf{v}_{k}$ over the interval $[k-1, k]$, and $\mathbf{w}_{i,k}$ is a term that accounts for process noise. 

The feature tracking measurement model for a single visual feature is expressed as
\begin{equation}
\mathbf{y}_{i,k}
= \mathbf{h}(\mathbf{X}_{i,k})+\mathbf{n}_{i,k},
\end{equation}
where $\mathbf{h}(\cdot)$ is the camera projection model and $\mathbf{n}_{i,k}$ denotes measurement noise. The expressions $\mathbf{f}(\cdot)$ and $\mathbf{h}(\cdot)$ are consistent with recursive Bayesian filtering, such as the EKF, which maintains the posterior probability of the state estimate

\begin{equation}
p_k = p(\mathbf{x}_k \mid \mathbf{y}_{1:k}),
\end{equation}
where $\mathbf{x}_k$ in this case corresponds to the relative 3-D position of a tracked landmark feature $\mathbf{X}_{i, k}$ and the perspective projection transformation of the feature to the image plane $\mathbf{y}_{i,k}$ is written more compactly as $\mathbf{y}_k$.

\subsubsection{Optimal Control Formulation for Active SLAM}
Active SLAM computes a control input that reduces the expected uncertainty of the future estimate. A general optimal control formulation considering a finite-horizon from step $0$ to $k_f$ is

\begin{equation}
\mathbf{u}_{0:k_f}^{\star}
=
\arg\min_{\mathbf{u}_{0:k_f}}
\mathbb{E}_{\mathbf{y}_{1:k_f}}
\left[ J(p_{k_f}) \right],
\end{equation}

where $J(\cdot)$ is the cost function. The posterior probability of the state estimate evolves according to the recursive Bayesian filtering model. The posterior covariance matrix $\mathbf{\Sigma}_{k}$ can be considered for the formalization of an estimation uncertainty metric to minimize such as \cite{Cristofalo2020_VisionBasedControl_3DReconstruction}:

\begin{equation}
J(p_k) = \operatorname{tr}(\mathbf{\Sigma}_k),
\end{equation}

Solving this active SLAM problem online can be computationally demanding. Hence, a one-step approximation can be adopted. Let

\begin{equation}
\hat{\mathbf{x}}_{k-1} = \mathbb{E}[\mathbf{x}_{k-1} \mid \mathbf{y}_{1:k-1}]
\end{equation}
denote the current state estimate, and let
\begin{equation}
\hat{\mathbf{x}}_{k}^{-}
=
\mathbf{f}(\hat{\mathbf{x}}_{k-1},\mathbf{u}_k)
\end{equation}

denote the predicted current state before incorporating the new measurement. A gradient-based law approximates $\mathbf{u}_k$ so that the state estimate can move along the direction of maximum uncertainty reduction:

\begin{equation}
\hat{\mathbf{x}}_{k}^{-}
=
\hat{\mathbf{x}}_{k-1}
-
\Gamma
\nabla_{\hat{\mathbf{x}}_{k-1}}
\mathbb{E}_{\mathbf{y}_{k}}
\left[ J(p_{k}) \mid
\mathbf{y}_{1:k-1}
\right],
\end{equation}
where $\Gamma$ is a control gain matrix. For posterior covariance minimization, this becomes \cite{Cristofalo2020_VisionBasedControl_3DReconstruction}:
\begin{equation}
\hat{\mathbf{x}}_{k}^{-}
=
\hat{\mathbf{x}}_{k-1}
-
\Gamma
\nabla_{\hat{\mathbf{x}}_{k-1}}
\operatorname{tr}
\left( \mathbf{\Sigma}_{k} \right),
\end{equation}

or equivalently

\begin{equation}
\mathbf{f}(\hat{\mathbf{x}}_{k-1},\mathbf{u}_k)
=
\hat{\mathbf{x}}_{k-1}
-
\Gamma
\nabla_{\hat{\mathbf{x}}_{k-1}}
\operatorname{tr}
\left(
\mathbf{\Sigma}_{k}
\right).
\end{equation}

By equating the desired evolution of the state estimates with the system dynamics and solving for the control input, the corresponding velocity command at step $k$ can be obtained as

\begin{equation}
\mathbf{v}_{k}
=
\frac{1}{dt}
\left(
\left( \mathbf{I}_{3} - e^{\mathbf{\Omega}_{k} dt} \right)\hat{\mathbf{X}}_{i, k-1}
-
\Gamma
\nabla_{\hat{\mathbf{X}}_{i, k-1}}
\operatorname{tr}
\left( \mathbf{\Sigma}_{k} \right)
\right).
\end{equation}

\subsubsection{Estimation Uncertainty Reduction for Optimal Beam Alignment}

The tracked UE or transmitter can be treated as a region of interest (ROI) in the image plane, retrieved using an object detection model \cite{osman2023vehiclecameras}. Within this ROI, the feature of interest can be tracked via the LK method. Let $\mathbf{X}_{\mathrm{tx},k} = [X_{\mathrm{tx},k}, Y_{\mathrm{tx},k}, Z_{\mathrm{tx},k}]^T$ denote its estimated 3-D position in the camera coordinate frame. The corresponding LOS path direction is:

\begin{equation}
\hat{\mathbf{d}}_k
= \frac{\mathbf{X}_{\mathrm{tx},k}}
{\left\|\mathbf{X}_{\mathrm{tx},k}\right\|_2}.
\end{equation}

The azimuth and elevation angles can then be obtained as

\begin{equation}
\phi_k = \operatorname{atan2}(Y_{\mathrm{tx},k},X_{\mathrm{tx},k}),
\end{equation}
\begin{equation}
\theta_k =
\operatorname{atan2}
\left(Z_{\mathrm{tx},k},
\sqrt{X_{\mathrm{tx},k}^2+Y_{\mathrm{tx},k}^2}
\right).
\end{equation}

Considering a transmitter / UE with a single omnidirectional antenna and a receiver beam-steering codebook $\mathcal{F}=\{\mathbf{f}_q\}_{q = 1}^{Q}$, the downlink received signal over the subcarrier $m$ can be modeled as \cite{osman2023vehiclecameras}:

\begin{equation}
z_{m, k}
=
\mathbf{h}_{m,k}(\phi_k, \theta_k)^{T}\mathbf{f}_{q, k} s_{k}
+
r_{m, k},
\end{equation}

where $\mathbf{h}_{m,k}(\phi_k, \theta_k)$ is the channel vector, $\mathbf{f}_{q, k}$ is the selected beamforming vector, $s_{k}$ is the transmitted symbol, and $r_{m, k}$ is receiver noise. The optimal beam that aligns better with $\hat{\mathbf{d}}_k$ yielding the strongest effective channel gain \cite{osman2023vehiclecameras}, can be selected as

\begin{equation}
q_k^{\star} = \arg\max_{q} \left| \mathbf{h}_{m,k}(\phi_k, \theta_k)^{T}\mathbf{f}_{q, k} \right|^2 .
\end{equation}

Fig. \ref{fig_7} visualizes the active SLAM framework described above. 

\begin{figure}[!t]
\centering
\includegraphics[width=\linewidth]{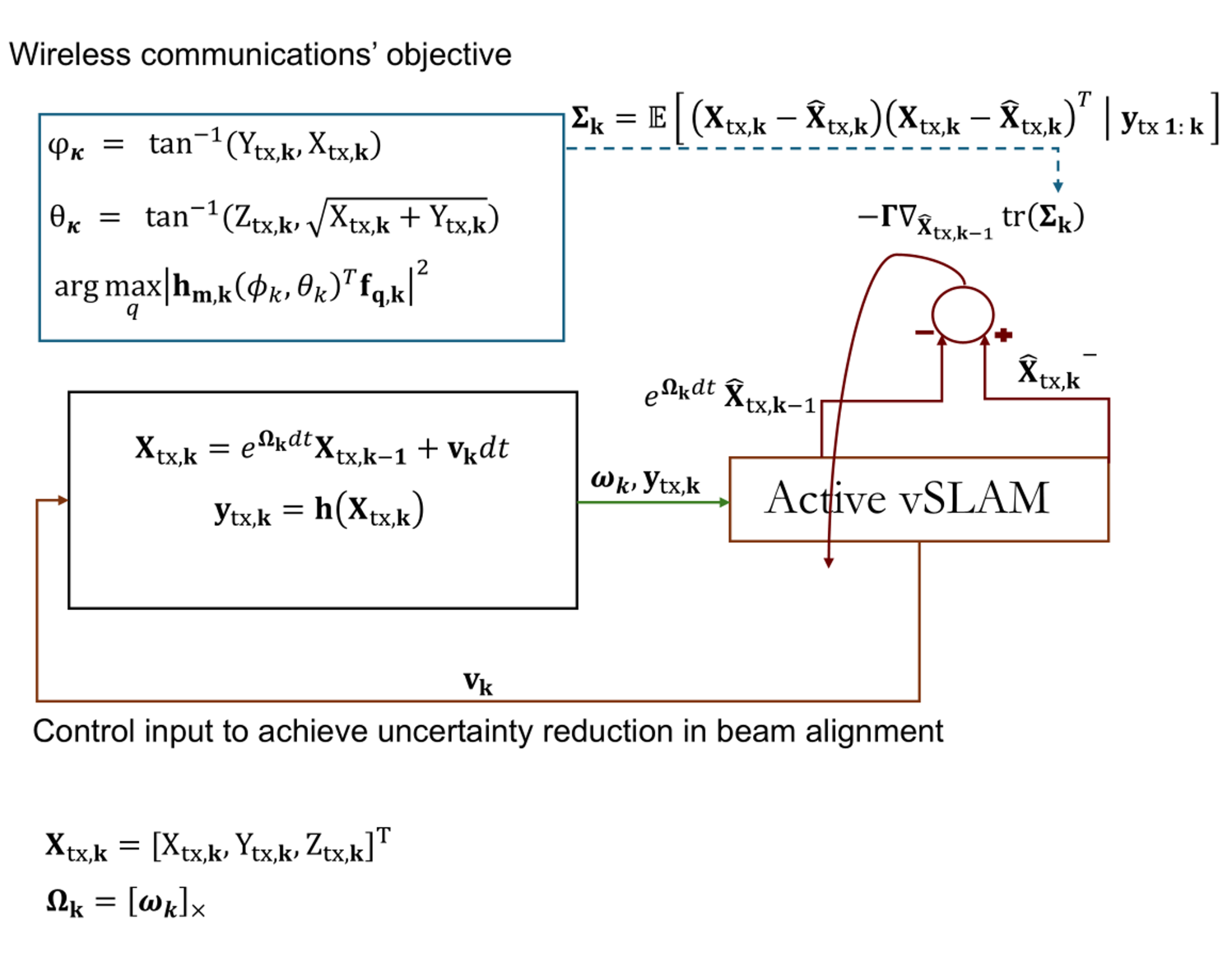}
\caption{{\color{blue}V-SLAM as an active framework that computes the next motion to achieve the communications objective of beam alignment through estimation uncertainty reduction $-\nabla_{\hat{\mathbf{X}}_{\mathrm{tx}, k-1}}
\operatorname{tr}
\left( \mathbf{\Sigma}_{k} \right)$ in the transmitter position $\hat{\mathbf{X}}_{\mathrm{tx},k}$.}}
\label{fig_7}
\end{figure}

\subsubsection{Limitations of the Active SLAM Uncertainty Reduction-driven Motion}
The above formulations demonstrate the capabilities of SLAM as an active estimation framework. Instead of estimating pose and landmarks from available vision and wireless communication measurements, the autonomous vehicle can perform auxiliary motions that could further reduce landmark uncertainty and assist wireless links over LOS paths. For aerial robots, $\mathbf{u}_k$ may directly represent a 3-D body-frame velocity command. For ground vehicles / cars, the same communications-driver active perception objective can be applied, however, $\mathbf{u}_k$ is restricted through nonholonomic constraints, longitudinal and lateral tire-slip dynamics, and constrained by lane, collision-avoidance, comfort and safety driving decisions.} {\color{blue}In addition to this}, having stopping criteria is imperative, because it can be shown that after reaching a particular point, more information would lead not only to a diminishing return effect but also, in case of ambiguous or contradictory information, to an unrecoverable state (e.g., numerous incorrect loop closures). Lastly, another critical research direction concerns the establishment of mathematical guarantees for active SLAM and the derivation of near-optimal control policies. Given that obtaining an exact, closed form solution to the problem is in the general case {\color{blue}computationally heavy}, developing approximation algorithms with well-defined performance bounds becomes essential. A representative example of this line of work is the application of submodularity theory \cite{Caillot2022_CoopPerceptionAutomotive} in the related domain of active sensor placement, which enables provable near-optimal solutions under certain conditions.

\section{SLAM Methods, Modalities and Techniques}

\subsection{Mathematical and Algorithmic Methods}

Mathematical modeling plays an important role in bridging SLAM with wireless communications. In conventional SLAM, probabilistic estimation techniques help robots and autonomous vehicles navigate and reconstruct their environment. However, when applied to wireless environments, SLAM must account for additional challenges posed by signal propagation, multipath effects, and dynamic radio environments. This {\color{blue}interaction} between localization, mapping, and wireless signal processing {\color{blue} motivates} mathematical frameworks that incorporate elements of RF-based sensor fusion, probabilistic inference, signal estimation, and information vectorization.

\subsubsection{Probabilistic Approaches in SLAM}

\begin{figure}[!t]
\centering
\includegraphics[width=\linewidth]{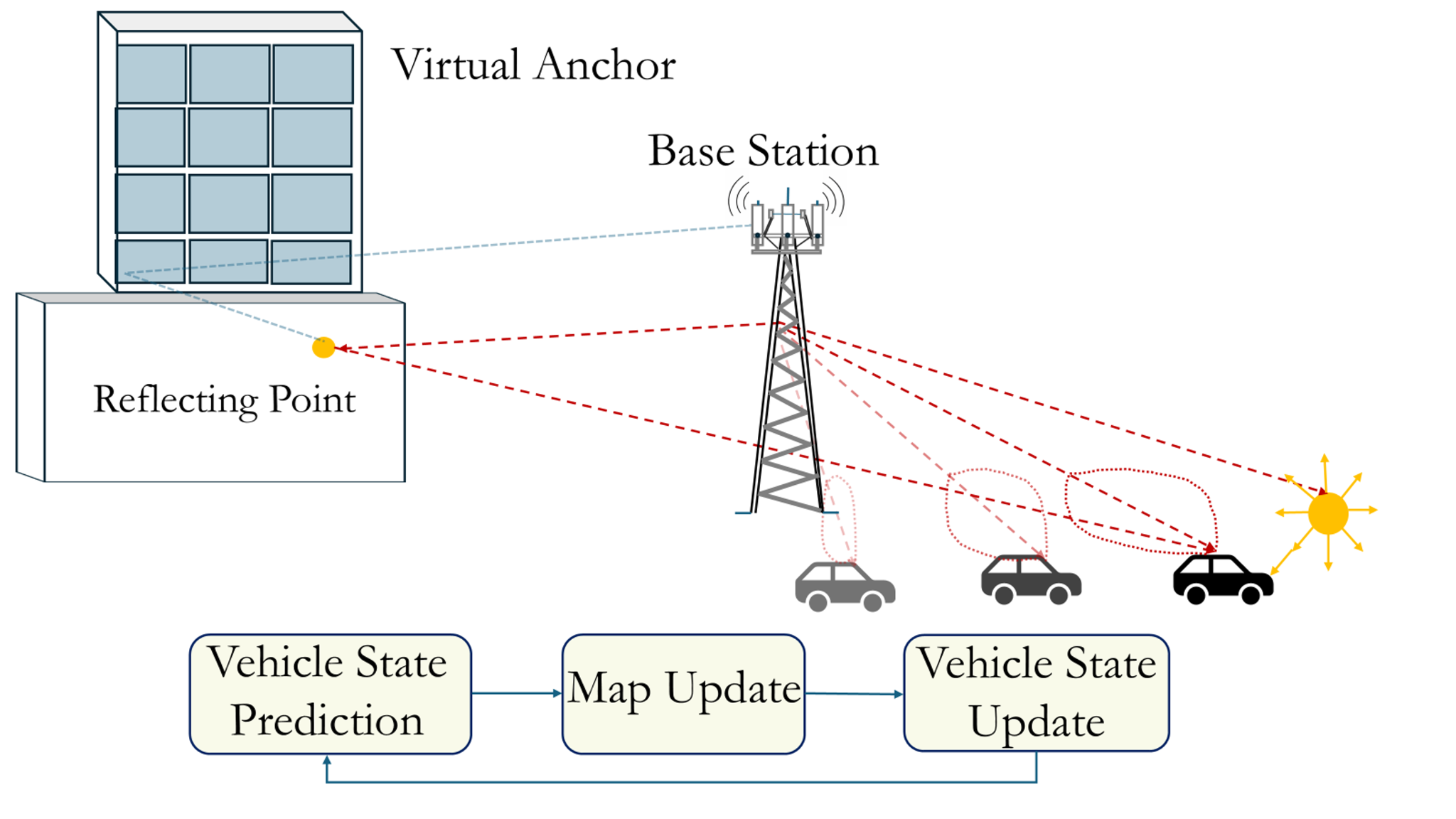}
\caption{Multi-dimensional channel estimation, channel parameter clustering and SLAM using a novel likelihood function that accounts for both specular and diffuse multipath components.}
\label{fig_8}
\end{figure}

In the topic of 5G \cite{Ge2020_5G_SLAM_ClusteringAssignment}, the authors investigated the localization and environment mapping capabilities of wireless systems operating above 24 GHz. An intermediate modeling approach was proposed, which differs from first-principles models and direct positioning methods. It consists of four phases: downlink data transmission, multidimensional channel estimation, channel parameter clustering, and SLAM, using a novel likelihood function that accounts for both specular and diffuse multipath components, as shown in Fig. \ref{fig_8}. A novel method was developed to cluster multipath components (MPCs) by projecting high-dimensional data into 3-D points, which are then clustered using an augmented density-based spatial clustering of applications with noise (DBSCAN) algorithm that accounts for channel gains.

Extending previous studies on 5G {\color{blue} positioning and environment mapping}, \cite{Barneto2022_mmWaveSensing_Mapping} explored localization techniques suitable for new radio (NR) and sub-THz 5G communications. They proposed an L1-regularized least-squares approach with low computational complexity to obtain sparse range-angle charts at individual sensing locations. In addition, a novel state model was introduced for tracking diffuse and specular scatterers using an interacting multiple model extended Kalman filter (IMM-EKF) and a smoother. The IMM design included a finite set of motion models with probabilistic transitions between them, following the principles of mmWave SLAM. Experimental results demonstrated the effectiveness of the proposed methods, with IMM-EKF-based tracking significantly outperforming static modeling approaches.

At the same pace as the exploration of mathematical solutions for RF-based localization, \cite{Giannelos2021_RFID_Localization_ParticleFiltering} focused on particle filters to facilitate localization based on phase measurements. The proposed methods introduced a weight metric for each particle-measurement pair based on geometric considerations and demonstrated robustness against phase measurement noise. The approach also estimated the unknown constant phase offset as a parameter without requiring reference tags or specific tag topology assumptions. Special attention was paid to the speed of algorithm execution, allowing real-time implementation on mobile robots equipped with RFID readers.

The integration of different sensor data modalities, while mathematically compensating for the uncertainties inherent in such synergies, has been considered in \cite{Yang2021_PlugPlay_Crowdsourcing_SLAM}. The aforementioned work extended the belief propagation (BP)-SLAM algorithm by integrating a mechanism for estimating time-varying agent locations, radio characteristics, measurement biases such as clock bias, IMU orientation bias, and RSS model parameters. This mechanism exhibits high accuracy and robustness even in challenging scenarios without prior knowledge of anchors and agents.

Focusing on the algorithmic aspects of their work, \cite{Kim2022_PMBM_SLAM_5GmmWave} proposed a sequence of three Poisson multi-Bernoulli mixture (PMBM)-based SLAM filters for ground vehicle-based SLAM. Their results showed that, compared to the probability hypothesis density SLAM filter \cite{Kim2022_PMBM_SLAM_5GmmWave}, these filters demonstrate superior robustness against missed detections and false alarms. The study also highlighted computational performance trade-offs arising from the marginalization of auxiliary variables and uncovered links between the PMBM and BP-SLAM filters, paving the way for complexity reduction and parallelized particle implementations to improve the accuracy of marginalized PMB-SLAM filters.

Among notable works in the area of SLAM mathematical methods, \cite{Wang2018_SLAM_RFID_ParticleFilters} presented a SLAM technique using independent particle filters for landmark mapping and localization in high-frequency band RFID systems. Independent particle filters estimate both the position and orientation of the robot and the positions of the integrated circuit (IC) tags. A comparative analysis with FastSLAM demonstrated the broader applicability of the proposed method, especially in cases where landmarks do not conform to Gaussian distributions. Simulated and real-world evaluations confirmed its superior performance over FastSLAM.

\subsubsection{Graph SLAM}

Similarly, using advanced graph algorithms, the authors of \cite{Yue2020_VisibleLight_GraphSLAM} presented a graph SLAM-based approach for indoor visible light map reconstruction. The method integrated pedestrian dead reckoning mathematically formalized constraints into a body motion pose graph. Visible light GraphSLAM improves both front- and back-end accuracy by refining trajectory estimation, which is then anchored to a floor map using a novel door detection technique. A visible light map was constructed by labeling RSS with corresponding trajectory locations, while a Kalman filter fuses visible light fingerprinting and inertial sensor data to improve user localization.

\subsubsection{Vector-field SLAM}

A more sophisticated approach based on vector calculus and difference equations was presented in \cite{Gutmann2014_VectorFieldSLAM_Extensions}. This study introduced a SLAM methodology that models spatial variations in continuous signals, such as WiFi RSS or beacon signals, for SLAM. The signals must be uniquely identifiable, stationary, and continuous in space. 

A vector field is defined as a regular grid with fixed node positions $\mathbf{b}_i$, where $i = \{1, \dots, N\}$, while each node $\mathbf{m}_i \in \mathbb{R}^M$ represents a vector of expected signal values when the robot is positioned at $\mathbf{b}_i$ with a fixed orientation $\theta_0$. Without loss of generality, let $\theta_0 = 0$. The vector field and robot pose are then {\color{blue}estimated} as follows.

Let the robot’s path be a time series of poses $\{ \mathbf{x}_t \}_{t=0}^{t=T}$, where for a given time instant $t$, $\mathbf{x}_t \in SE(2)$ (the set of rigid body transformations in the two-dimensional horizontal plane). Therefore, it is convenient to assume that the planar robot kinematics evolve in a manner that is purely determined by:

\begin{equation}
\mathbf{x}_t =
\begin{bmatrix}
x(t) \\
y(t) \\
\theta(t)
\end{bmatrix}.
\end{equation}

Moreover, let the initial pose be
\begin{equation}
\mathbf{x}_0 = \begin{bmatrix} 0 \\ 0 \\ 0 \end{bmatrix}.
\end{equation}

At each time step $t = \{1, \dots, T\}$, the robot executes a motion/control input $\mathbf{u}_t$ with covariance $\mathbf{R}_t$ and receives a measurement $\mathbf{z}_t$ of the continuous robot state signals with covariance $\mathbf{Q}_t$.

The motion of the robot is estimated by a nonlinear state-space model
\begin{equation}
\mathbf{x}_t = \mathbf{g}(\mathbf{x}_{t-1}, \mathbf{u}_t) + \mathbf{e}_u ,
\end{equation}

where $\mathbf{e}_u$ is a zero mean error with covariance $\mathbf{R}_t$.  

\begin{figure}[!t]
\centering
\includegraphics[width=\linewidth]{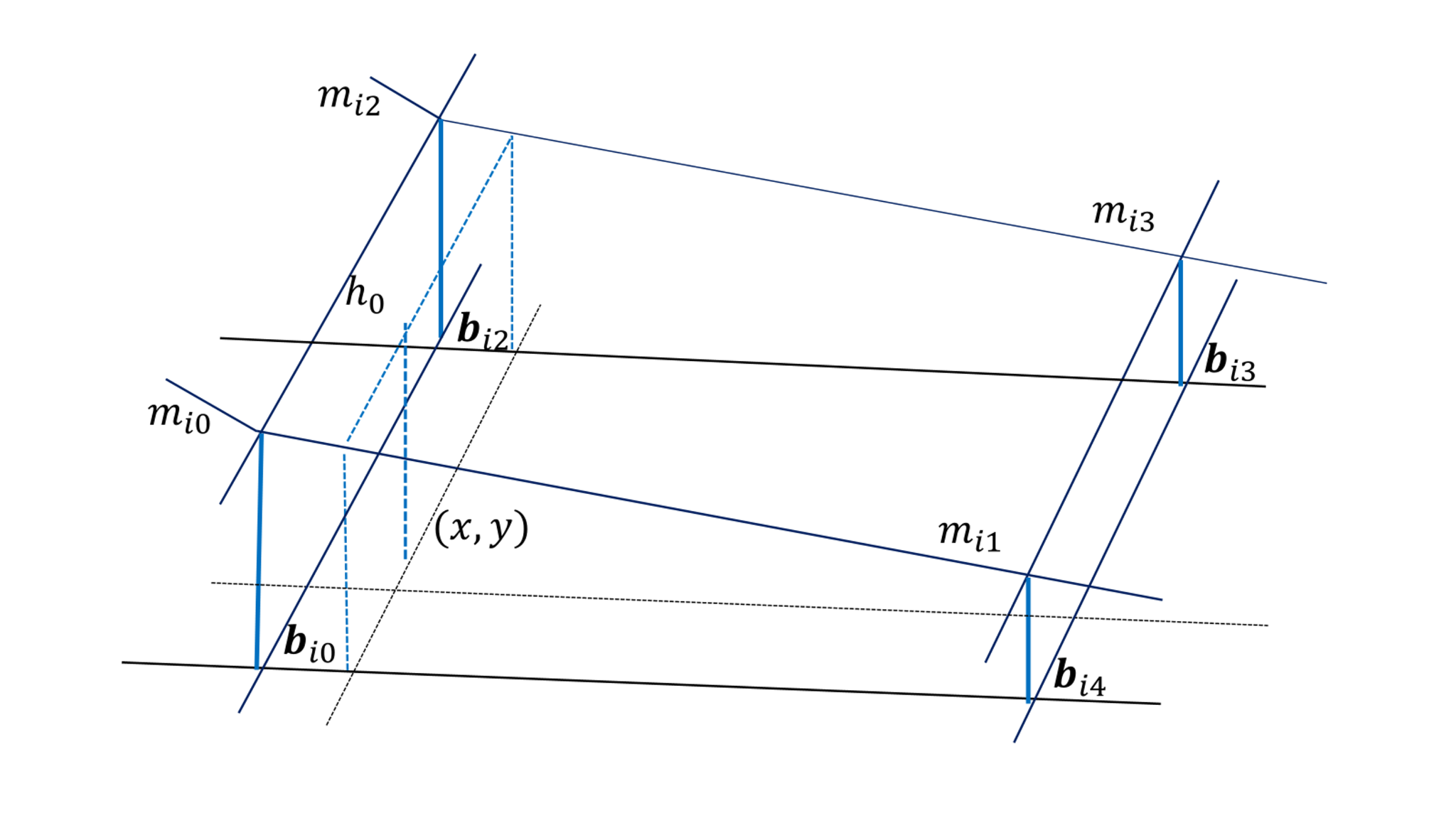}
\caption{Vector Field SLAM: The bilinear interpolation function of the expected signal values from the four nodes of the cell containing the robot, based on \cite{Gutmann2014_VectorFieldSLAM_Extensions}.}
\label{fig_9}
\end{figure}

Additionally, the measurement model outputs an observation given the current robot pose and the vector field

\begin{equation}
\mathbf{z}_t = \mathbf{h}\left(\mathbf{x}_t, \begin{bmatrix} \mathbf{m}_1, & \dots, & \mathbf{m}_N \end{bmatrix}^T \right) + \mathbf{e}_z ,
\end{equation}

where $\mathbf{e}_z$ is a zero mean error with covariance $\mathbf{Q}_t$. The considered measurement model consists of a rotational and a translational component:

\begin{equation}
\begin{aligned}
h\!\left(\mathbf{x}_t,
[\mathbf{m}_1,\ldots,\mathbf{m}_N]^T\right)
=
h_R\!\left(
h_0\!\left(x,y,
[\mathbf{m}_1,\ldots,\mathbf{m}_N]^T\right),
\theta
\right),
\end{aligned}
\label{eq:vslam_observation_model}
\end{equation}

where $h_R$ is a continuous function that rotates the expected signal values based on the orientation of the robot $\theta$, and $h_0$ is a bilinear interpolation function of the expected signal values of the four nodes of the cell containing the robot, as also illustrated in Fig. \ref{fig_9}:

\begin{equation}
h_0 \left(x, y, \begin{bmatrix} \mathbf{m}_1, & \dots, & \mathbf{m}_N \end{bmatrix}^T \right) = \sum_{j=0}^{3} w_j \mathbf{m}_{i_j} .
\end{equation}


{\color{blue}Table \ref{tab:slam_methods_comparison} summarizes the key characteristics of the discussed mathematical and algorithmic methods.}

\begin{table*}[t]
\centering
{\color{blue}
\caption{Comparison of Representative Mathematical and Algorithmic SLAM Methods}
\label{tab:slam_methods_comparison}
\scriptsize
\setlength{\tabcolsep}{3pt}
\renewcommand{\arraystretch}{1.18}
\begin{tabular}{|
>{\raggedright\arraybackslash}p{2.6cm}|
>{\raggedright\arraybackslash}p{3.0cm}|
>{\raggedright\arraybackslash}p{3.0cm}|
>{\raggedright\arraybackslash}p{2.8cm}|
>{\raggedright\arraybackslash}p{2.8cm}|
>{\raggedright\arraybackslash}p{2.2cm}|}
\hline
\textbf{SLAM Method Category} 
& \textbf{Key Characteristics} 
& \textbf{Application Domain / Signal Modalities} 
& \textbf{Advantages} 
& \textbf{Constraints} 
& \textbf{Representative Works} \\
\hline

\textbf{Probabilistic Approaches in SLAM} 
&
Likelihood-based modeling accounting for multipath effects (specular/diffuse); multi-model estimation (filter banks, particle filtering); recursive Bayesian inference.
&
5G/mmWave positioning; RFID-based localization; inertial \& RSS fusion.
&
High accuracy and robustness; relaxation of Gaussianity assumption; robust under limited prior information.
&
High computational complexity; particle degeneracy; scalability challenges in dense environments.
&
\cite{Ge2020_5G_SLAM_ClusteringAssignment,Barneto2022_mmWaveSensing_Mapping,Giannelos2021_RFID_Localization_ParticleFiltering,Yang2021_PlugPlay_Crowdsourcing_SLAM,Kim2022_PMBM_SLAM_5GmmWave,Wang2018_SLAM_RFID_ParticleFilters}
\\
\hline

\textbf{Graph-based SLAM} 
&
Factor graph formulation for global optimization; pose graph construction from motion constraints; trajectory refinement via optimization.
&
Visible light positioning; monocular/stereo SLAM; RSS-based localization; fusion with inertial data.
&
Improves global map consistency; reduces drift via loop closure; supports large-scale mapping.
&
Computational overhead in large graphs; sensitivity to incorrect loop closure.
&
\cite{Yue2020_VisibleLight_GraphSLAM}
\\
\hline

\textbf{Vector Field SLAM} 
&
Continuous spatial signal field representation; RF signals modeled as vector-valued fields; interpolation-based observation models.
&
WiFi RSS-based positioning; RF-based SLAM; structured light pattern-based positioning.
&
Suitable for signal-based mapping; no explicit landmark extraction required.
&
Assumes signal stationarity; requires identifiable signals; sensitive to dynamic environments.
&
\cite{Gutmann2014_VectorFieldSLAM_Extensions}
\\
\hline

\end{tabular}}
\end{table*}

\par
{\color{blue}
\textit{Lessons Learned:}
\begin{enumerate}
    \item \textit{Modeling Fidelity:} Likelihood-based formulations can enable the incorporation of the complex scattering nature of RF propagation, yielding high accuracy and robustness. Factor graph representations can naturally encode geometric and motion constraints in visual and inertial systems, resulting in enhanced global map consistency and reduced drift in pose estimation. Vector field formulations enable high-fidelity signal-driven modeling without the need for explicit determination of landmarks.
    \item \textit{Complexity:} Probabilistic and graph-based approaches can incur significant computational cost in large-scale, dense environments. Vector field methods alleviate the need for explicit modeling through functional approximations, at the expense of stronger assumptions made on signal stationarity and identifiability.
\end{enumerate}
}

{\color{blue}\subsection{Sensor Modality-Driven SLAM Techniques}}

\subsubsection{Inertial SLAM}
Miniature inertial measurement units (MIMU) have gained significant popularity in various applications, with inertial odometry for pedestrian navigation systems (PNS) being a notable example. The authors of \cite{Li2024_3DCOGI_SLAM_SFE} proposed a 3-D collaborative occupancy grid-based inertial SLAM solution (3D-COGI-SLAM), which utilizes switching floor event matching. This approach aims to mitigate the challenges associated with position error divergence, instability of range information, and limited access to initial relative states in MIMU-PNS. Despite the advances in MIMU-based localization, further improvements in orientation estimation remain critical to improving overall system performance. Building on the role of inertial SLAM, the authors of \cite{Que2023_JointBeamManagement_SLAM} presented a framework that estimates path angles through a hierarchical sweep approach. This method enables angle-based localization and mapping by utilizing an IMU to support improved sensing accuracy. The integration of IMU-based angle estimation contributes to improved localization reliability, addressing a key challenge in scenarios where other sensing modalities may be insufficient.

While inertial and angle-based SLAM offer promising solutions, point-line-based visual inertial SLAM remains essential, especially for mobile robots operating in dynamic environments. However, traditional point-line-based V-SLAM systems often suffer from localization inaccuracies and frequent tracking losses, especially in complex urban environments. As a result, hybrid approaches that combine the strengths of multiple modalities have received increasing attention from the robotics community.  To overcome such limitations, the authors of \cite{Kuang2022_MonocularVISLAM_6G} proposed a real-time, robust point-line-based monocular visual inertial SLAM (VINS) system, tailored for smart cities and future 6G networks. By employing Edge Drawing Lines with adaptive gamma correction, the system effectively extracts a higher proportion of long-line features, improving feature robustness in challenging scenes. In addition, a real-time line feature matching method is introduced, which enables efficient tracking of extracted line features between adjacent frames without relying on descriptor computation. In the front-end, spatial lines are reconstructed through feature tracking, while in the back-end, camera states are refined by minimizing a cost function that incorporates point-line reprojection errors and IMU residuals.

\subsubsection{LiDAR-based SLAM}
LiDAR-based SLAM is widely used for positioning and navigation. By projecting a series of light pulses onto the surface of an object, information about its relative distance and pose is obtained. In the spirit of fusing LiDAR and optical sensory information with GNSS signals, the authors of \cite{He2024_LiDARSLAM_GNSSGraphOpt} addressed the challenge of improving the accuracy of pose optimization without directly affecting position estimation. Non-RF GNSS-based pose optimization algorithms rely on alternative information sources, such as visual features. However, factors such as lighting conditions, occlusions, and calibration errors in optical sensors and cameras can degrade accuracy. Integrating GNSS with pose optimization algorithms can significantly improve performance compared to non-GNSS approaches. The study in \cite{He2024_LiDARSLAM_GNSSGraphOpt} combined a graph optimization algorithm with GNSS to refine LiDAR SLAM pose estimation. By using this graph optimization framework, the GNSS-based LiDAR SLAM pose optimization model achieved higher accuracy in pose estimation. The optimal pose was derived by evaluating constraints within the defined space. However, due to dynamic environmental variations and uncertainty factors that fluctuate over time, obtaining a truly optimal pose remains a challenge. The study demonstrated that incorporating GNSS pose optimization results in a \%99 reduction in distance, level, and height deviations compared to non-GNSS approaches.

\begin{figure}[!t]
\centering
\includegraphics[width=\linewidth]{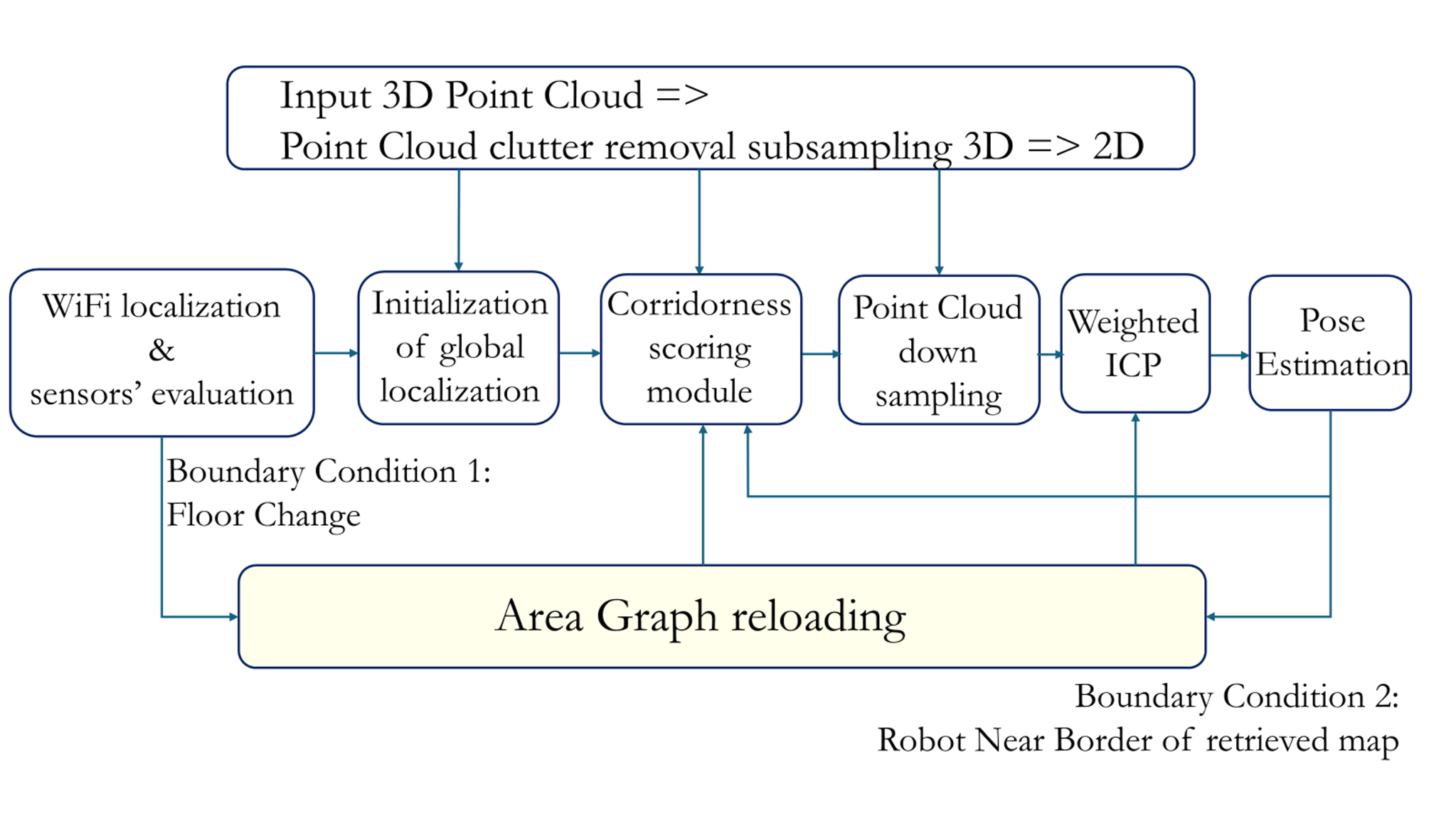}
\caption{Area graph localization (AGloc) pipeline based on \cite{Xie2023_LifelongLiDAR_Localization_AreaGraph}.}
\label{fig_10}
\end{figure}

In \cite{Bi2024_HW_SW_CoDesign_InspectionRobot}, the authors investigated hardware and software co-design strategies to meet the requirements of mobile robot localization and object perception. To enable accurate mobile positioning, they proposed a SLAM strategy that integrates LiDAR, a monocular camera, and an IMU. For object perception, the study exemplifies the extraction of meter information using DL techniques for meter detection, localization, and text recognition. The integration of 5G and WiFi communication enables an intelligent robotic inspection system with a network management platform. The authors of \cite{Xie2023_LifelongLiDAR_Localization_AreaGraph} addressed the challenge of robust long-term localization in cluttered indoor environments, such as offices and corridors, using LiDAR point clouds within an area graph. This hierarchical, topometric semantic map representation uses polygons to define spaces such as rooms, corridors, or buildings. Their approach, called area graph localization (AGloc), is augmented by a clutter removal subsampling method to remove excessive noise, as well as sampling techniques and weighting functions for global robot localization (Fig. \ref{fig_10}). First, a set of candidate robot positions is generated using a sampling technique, after which a scoring function evaluates each candidate based on its match with the area graph. The point-to-point iterative closest point (ICP) algorithm and a weighting function enable localization using only wall and passage representations, ensuring robust lifelong localization despite the presence of non-static obstacles such as furniture, pets, and humans. Additionally, a 'corridorness' score function is introduced to provide downsampled point clouds, improving ICP localization in corridor-like environments.

Although numerous studies have investigated LiDAR-based SLAM, it is valuable to examine the underlying technology of LiDAR, specifically laser range finding (LRF). In \cite{Kim2022_ElectricFieldSim_2DMapSLAM}, the authors investigated the kidnapped robot scenario, where RF/mmWave-based localization is unavailable due to disrupted wireless communication. The proposed system allows a mobile robot to scan an unknown environment using an LRF sensor. However, the study found that LRF data often contains errors, resulting in ragged point clouds even when smooth surfaces are detected. To mitigate this problem, an information fusion approach was adopted. The generated line segment map was enriched with additional information such as material properties and thickness. By integrating these data, the study demonstrated that the electric field strength can be effectively simulated using the finite-difference time-domain method.

In \cite{Zhou2022_MultiFloor_Localization_MultiModal}, a high-precision multimodal approach for indoor positioning was proposed using LiDAR as the central sensor modality. This method improves localization accuracy by incorporating widely available technologies, such as WiFi access points (APs) and low-cost cameras. In this hybrid scheme, positioning data from LiDAR-based SLAM is used as a theoretical reference for visual images, while WiFi signals estimate the rough location through likelihood. Finally, RGB images refine the ground map and improve the localization accuracy. As mentioned above, LiDAR-based localization is prone to errors, especially due to environmental changes and illumination variations. To provide a reasonable basis for comparing 5G/mmWave sensing and LiDAR perception capabilities, the authors of \cite{Karfakis2023_NR5GSAM} proposed a cellular SLAM framework that leverages 5G NR signals and inertial measurements for mobile robot localization using multiple BSs. Their method estimates the robot's pose while generating a radio signal map based on RSSI measurements for correction purposes. In the study, this approach was benchmarked against LiDAR-inertial odometry smoothing and mapping by comparing its performance against ground-truth references in simulations. Two experimental setups were evaluated using sub-6 GHz and mmWave frequency bands for communication, with transmission based on downlink signals. The results showed that 5G positioning can be effectively used for radio SLAM, improving robustness in outdoor environments and serving as an additional absolute reference source when LiDAR-based methods fail.

\subsubsection{Visual Range Odometry-based SLAM}
Significant efforts have been made to achieve accurate real-time navigation using SLAM and pose estimation in environments and applications where high accuracy can practically serve an application scope. An indoor wayfinding system based on geometric features-aided graph SLAM for the visually impaired has been presented in \cite{Zhang2017_GeometricGraphSLAM_IndoorWayfinding}. The authors of this study addressed the challenge of indoor navigation within the framework of a robotic navigation aid, proposing a client-server architecture for real-time wayfinding. Producing both range data and intensity images at rates of up to 50 FPS, the SR4000 is well-suited for SLAM applications due to its compact dimensions (50 × 48 × 65 mm³) and configurable modulation frequency, allowing multiple cameras to operate simultaneously without interference. In terms of architecture, the client device captures data from the camera and transmits it to the server via WiFi while also interacting with the user through a speech interface. The server, in turn, performs SLAM, localizes the user within a floor plan, plans the optimal path, and sends navigational commands and points-of-interest messages back to the client. To estimate the camera’s egomotion, the system employed the visual range odometry (VRO) algorithm \cite{Nister2004_VisualOdometry}, which extracts visual features from the current intensity image and matches them with those in the previous frame. The SIFT feature descriptors \cite{Lowe2004_SIFT_Features} facilitate feature matching, while siftGPU \cite{Sinha2006_GPU_VideoFeatureTracking} accelerates feature extraction and matching, achieving a runtime of approximately 30 milliseconds for VRO. The SLAM approach follows a two-step process: first, it extracts the floor plane from the camera’s point cloud data and integrates it as a landmark in a 3-D graph SLAM process, reducing pitch, roll, and Z-axis errors. Next, it projects the point data onto the extracted floor plane to detect wall lines, incorporating them into a 2D graph SLAM (in the OXY plane) to minimize yaw, X, and Y errors. By leveraging geometric features-namely, floor planes and wall lines-this two-step SLAM method effectively reduced 6-DOF pose estimation errors.

\subsubsection{RSSI-based Multi-agent Localization}
Multi-agent localization in wireless environments is another important topic. Koutsonikolas et al. \cite{Koutsonikolas2006_CoCoA_CooperativeLocalization} proposed CoCoA, an architecture for low-cost, energy-efficient localization in dynamic, infrastructure-less environments using cooperative mobile robot teams. The key component of CoCoA is a beacon-based localization scheme that uses RSSI for ranging and Bayesian inference to estimate node positions. It incorporates a motion-odometry model where velocity is chosen within predefined limits that vary across simulation experiments. The system achieved an average localization error of about 8 m when only one-third of the robots are equipped with localization devices, making it particularly useful for applications such as search and rescue. 

{\color{blue}Table \ref{tab:slam_modalities_overview} presents an overview of the discussed SLAM modality-based techniques, highlighting their key characteristics.}

\begin{table*}[t]
\centering
{\color{blue}
\caption{Overview of SLAM Modality-Based Techniques and Core Characteristics}
\label{tab:slam_modalities_overview}
\scriptsize
\setlength{\tabcolsep}{3pt}
\renewcommand{\arraystretch}{1.18}
\begin{tabular}{|
>{\raggedright\arraybackslash}p{3.0cm}|
>{\raggedright\arraybackslash}p{5.0cm}|
>{\raggedright\arraybackslash}p{3.5cm}|
>{\raggedright\arraybackslash}p{2.5cm}|}
\hline
\textbf{SLAM Modality} 
& \textbf{Key Characteristics} 
& \textbf{Core Sensing Technologies \& Algorithms} 
& \textbf{Representative Studies} \\
\hline

\textbf{Inertial SLAM} 
&
3-D collaborative occupancy grid-based inertial SLAM; switching floor event matching; effective motion estimation under limited external metric references.
&
MIMU; pedestrian navigation systems (PNS); visual-inertial navigation systems (VINS).
&
\cite{Li2024_3DCOGI_SLAM_SFE,Que2023_JointBeamManagement_SLAM,Kuang2022_MonocularVISLAM_6G}
\\
\hline

\textbf{LiDAR-based SLAM} 
&
Point cloud mapping extendable by GNSS and inertial sensing; improving indirectly the accuracy of pose optimization; robust long-term localization in structured environments within area graphs (AGLoc); augmenting wireless-enabled frameworks utilizing WiFi APs.
&
LRF; LiDAR-extended sensor fusion; AGLoc; LiDAR \& WiFi \& camera-based fusion; ICP.
&
\cite{He2024_LiDARSLAM_GNSSGraphOpt,Bi2024_HW_SW_CoDesign_InspectionRobot,Xie2023_LifelongLiDAR_Localization_AreaGraph, Kim2022_ElectricFieldSim_2DMapSLAM, Zhou2022_MultiFloor_Localization_MultiModal, Karfakis2023_NR5GSAM}
\\
\hline

\textbf{Visual Range Odometry-based SLAM} 
&
Visual range odometry using feature extraction and matching; GPU-accelerated SIFT descriptors; real-time pose estimation and mapping.
&
3-D cameras; SIFT feature descriptors; GPU acceleration; visual odometry.
&
\cite{Zhang2017_GeometricGraphSLAM_IndoorWayfinding,Nister2004_VisualOdometry,Lowe2004_SIFT_Features,Sinha2006_GPU_VideoFeatureTracking}
\\
\hline

\textbf{Multi-agent RSSI-based Localization} 
&
Beacon-based cooperative localization using RSSI measurements; Bayesian inference; multi-robot coordination.
&
RSSI sensing; wireless beacons; multi-robot systems; cooperative localization frameworks.
&
\cite{Koutsonikolas2006_CoCoA_CooperativeLocalization}
\\
\hline

\end{tabular}}
\end{table*}

\par
{\color{blue}
\textit{Lessons Learned:} 
\begin{enumerate}
    \item Inertial SLAM can serve as an alternative to V-SLAM in indoor environments, enabling 3-D occupancy grid reconstruction and motion estimation, including handling of switching floor events. LiDAR-based SLAM naturally provides dense 3-D point cloud representations, which, combined with algorithms such as DBSCAN and ICP, enable robust tracking of structural entities. Often integrated within sensor fusion frameworks, LiDAR combined with visual, inertial, and wireless systems can further enhance performance. Visual range odometry-based SLAM, particularly when GPU-accelerated, can support accurate real-time pose estimation and mapping. In view of cooperative localization, multi-agent systems can provide state estimates using RSSI measurements and relying on Bayesian estimation frameworks.
\end{enumerate}
}

{\color{blue}\section{Wireless- and 6G-Enabled SLAM}}

{\color{blue}Wireless communications in the era of beyond-5G are considered a key enabler of AI and especially ML/DL frameworks that aim to perform in real-time. Whilst the high transmission data rates and the effective retrieval of deep embeddings sent over wireless links provide the capabilities to support autonomous systems' applications, properties at the physical layer, through the utilized wider spectrum of high frequencies, can be particularly important for localization and mapping. In this section, radio waves are considered as SLAM-related input features that form high-dimensional spaces, resulting in the learning of representations in a manner analogous to computer vision ML/DL embeddings.}

\subsection{DNN-based Wireless SLAM}
DNNs have emerged as a transformational enabler for wireless SLAM, allowing the fusion of high-dimensional sensory data with wireless communications-based localization cues. Unlike traditional probabilistic or optimization-based SLAM pipelines, DNNs can learn complex nonlinear mappings between radio features and spatial representations, significantly enhancing robustness under multipath and NLOS conditions. These learning-driven frameworks can also be capable of real-time adaptability to dynamic environments and infrastructure changes, with the latter being an essential requirement in 6G and beyond wireless systems. Two representative DNN based wireless SLAM paradigms have recently gained prominence: deep coupling (DC)-SLAM, which tightly integrates wireless channel parameters with kinematic states for robust localization, and deep semantic and spatial joint perception, which extends SLAM toward semantic scene understanding and dense 3-D reconstruction. Together, these paradigms illustrate the convergence of DL, sensing, and wireless communications technologies in the realization of intelligent, environment-aware perception systems.

\subsubsection{Deep Coupling SLAM}
Addressing indoor positioning challenges, particularly in scenarios affected by NLOS interference where conventional 5G positioning techniques such as triangulation and trilateration struggle with precision, the authors of \cite{Sun2024_DC_SLAM_MAPPF} introduced the DC-SLAM method. This approach integrates multiple parameters, including delay, angle, and power amplitude of the observed CSI, alongside system state information such as the location, velocity, and virtual anchor position of the UE. Additionally, DC-SLAM tackles the challenge of multipath birth and death, enabling adaptation to dynamic environments. In experiments conducted in a typical indoor office setting, the maximum a posteriori-penalty Function achieved an average positioning accuracy of 0.11 meters under identical conditions, significantly outperforming the BP-SLAM method, which achieved only 0.48 meters.

\subsubsection{Deep Semantic and Spatial Joint Perception}
Extending the discussion to the technical implications of joint spatial 3-D perception and neural metric-semantic understanding in the context of 5G \& beyond edge computing, the authors of \cite{Zhu2021_SemanticSpatial_JointPerception} highlight Dense Semantic 3-D Reconstruction as a key component of this integrated approach. The general pipeline of metric-semantic understanding leverages depth maps and pixel-wise semantic classification scores to achieve the ultimate goal of semantic understanding in 3-D environments. In dense semantic 3-D reconstruction, class likelihoods corresponding to semantic masks are fused with depth masks using a convex optimization paradigm, resulting in a dense 3-D model of the reconstructed scene.

\subsection{Radio Maps Extraction Algorithm via mmWave Sensing}
Radio map extraction via RF sensing in indoor environments is a critical aspect of real-time and accurate mapping in SLAM. The challenges become even more pronounced in public safety scenarios, where infrastructure may be unreliable or infeasible to deploy, and the environment is highly dynamic due to varying ambient conditions, uncontrolled movements, and other factors. Additionally, traditional SLAM methods have inherent limitations tied to the sensing modalities they rely on \cite{Basu2024_UbiquitousIndoorMapping_RadioTomography}. Standard SLAM techniques, utilizing LiDAR, acoustic sensors, mmWave, and monocular or stereo camera-based approaches, are highly precise but constrained to detect structures within the direct LOS. This often necessitates extended scanning trajectories and provides additional structural details such as thickness and material properties.

To overcome these limitations, radio tomographic imaging (RTI) has been explored for mapping beyond LOS. Unlike traditional SLAM modalities, RTI leverages RF, typically within the sub-6 GHz spectrum, allowing signals to penetrate building structures and capture information beyond visual obstructions. Given a signal attenuation image, standard wireless fading models can estimate total propagation loss between two points on a map, referred to as the forward problem. However, RTI aims to solve the inverse problem, where propagation loss measurements from localized transceivers are used to estimate the attenuation image, thereby reconstructing the structural layout.

In this context, the authors of \cite{Basu2024_UbiquitousIndoorMapping_RadioTomography} proposed UBIQMAP, a lightweight, RTI-based end-to-end system for real-time indoor mapping with minimal to zero reliance on pre-deployed infrastructure. UBIQMAP was tested in various settings, including real-world deployments in a moderately complex residential apartment (800 sq. ft) and a large building foyer (278 sq. meters), as well as in multiple simulated scenarios. The study demonstrates the advantages of UBIQMAP over traditional SLAM techniques in specific contexts and advocates for integrating RTI with SLAM to enhance future mapping technologies.

Extending the discussion to mmWave networks, the authors of \cite{RastorguevaFoi2024_mmWaveRadioSLAM} presented an end-to-end processing framework focusing on channel parameter estimation and landmark extraction from raw in-phase/quadrature (I/Q) signal data. This approach addresses the emerging challenge of cellular bistatic SLAM in mmWave networks. A key contribution is a novel singular value decomposition (SVD)-based estimation method to derive AoA and angle of departure (AoD) parameters for propagation paths. The proposed method operates on beam reference signal received power (BRSRP) measurements without requiring detailed knowledge of complex antenna patterns, steering vectors, or beamforming weights. Notably, it incorporates built-in robustness against antenna side lobes by leveraging the natural sparsity of the beamformed mmWave channel. Additionally, the study introduces a new SLAM technique for jointly estimating landmark and UE positions, achieving superior robustness and identifiability compared to prior approaches. The proposed methods were rigorously evaluated using ray tracing and real-world measurement data at 60 GHz.

\begin{figure}[!t]
\centering
\includegraphics[width=\linewidth]{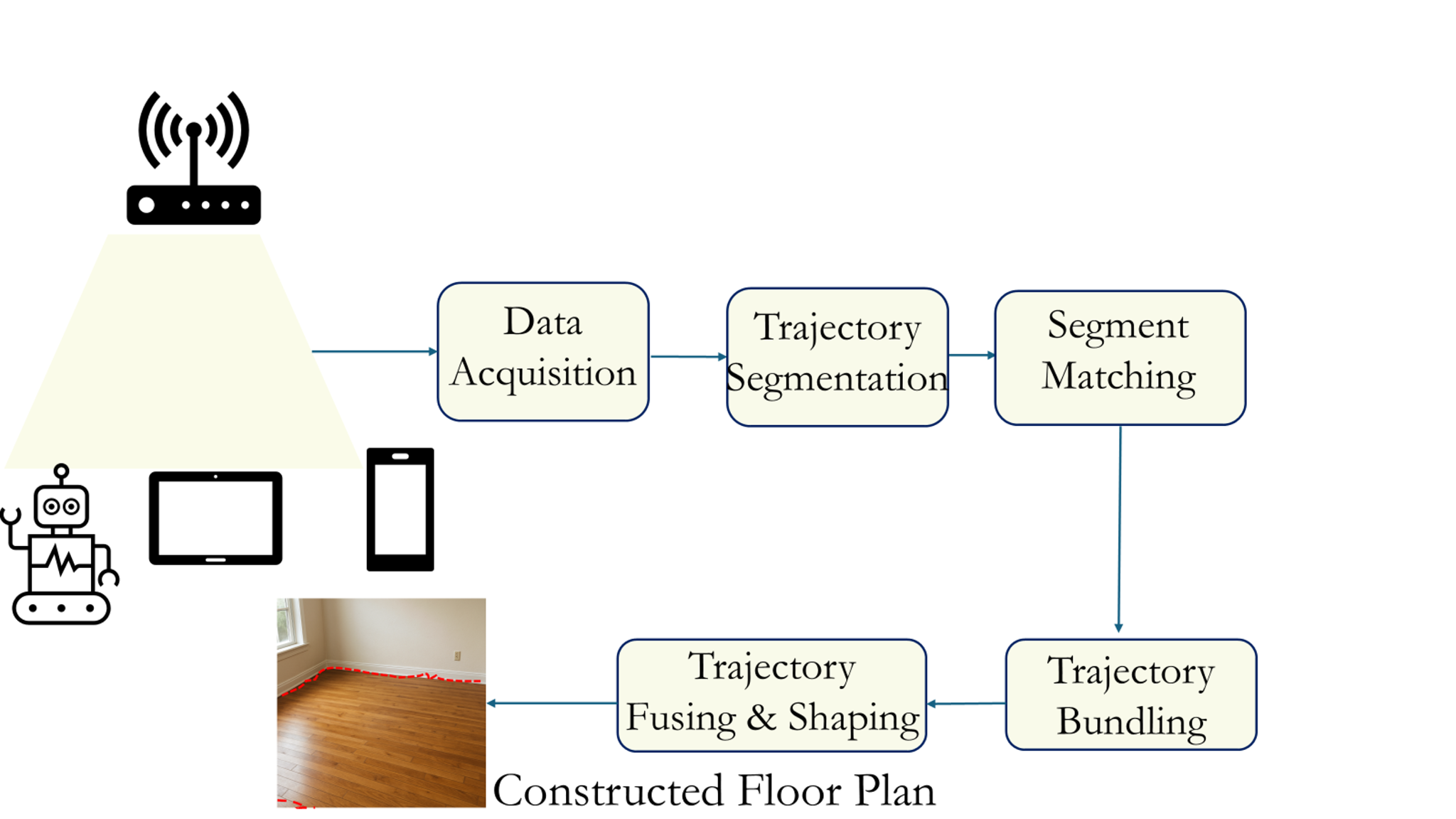}
\caption{EZMAP: a high-accuracy, low-cost floor plan construction system that fuses RF tracking with inertial sensing, based on \cite{Zhu2023_EZMap_FloorPlanConstruction}.}
\label{fig_11}
\end{figure}

The study in \cite{Que2023_JointBeamManagement_SLAM} addressed the significant overhead imposed by mmWave beam management in SLAM implementations that rely on AoA/AoD angle estimates. The authors proposed a joint beam management and SLAM framework that exploits the strong coupling between the radio map and multipath propagation for simultaneous beam management, localization, and mapping. For localization purposes, beam-sweeping-derived sensory data are fused with kinematic information from an IMU. Furthermore, the authors introduced a feature-assisted tracking method that incorporates prior angle information extracted from the radio map and IMU data. To enhance UE position and velocity estimates, a second-order discrete-time system was employed. Further advancing RF-based inertial tracking, the authors of \cite{Zhu2023_EZMap_FloorPlanConstruction} introduced EZMAP, a high-accuracy, low-cost floor plan construction system that fuses RF tracking with inertial sensing, as shown in Fig. \ref{fig_11}. EZMAP effectively combines the fine-grained yet localized RF information with the coarse-grained but global inertial sensing data (e.g., magnetic field strength), yielding an accurate and scalable mapping solution.

The work in \cite{Li2023_CrowdFusion_VisibleLightSLAM} introduced crowd fusion SLAM, a coarse-grained and fine-grained trace merging framework based on iterative closest point (ICP) SLAM and graph SLAM. This approach leverages visible light sensing for accurate loop closure detection while incorporating magnetic fields to extend mapping coverage. The key contributions of this study include: (i) an enhanced signal SLAM introducing a two-step trace merging process integrating ICP-based alignment, (ii) robust loop closures through visible light-based detections and (iii) an improved particle filtering method applied after trace merging to construct radio maps enriched with visible light and WiFi fingerprinting.

The study in \cite{Lotti2023_RadioSLAM_THz_6G} explored the challenges posed by next-generation wireless communication, where user devices autonomously scan their surroundings by steering narrow antenna beams and collecting echoes from objects and walls. The authors developed radio-SLAM algorithms inspired by image signal processing techniques, leveraging mobile THz radar for environment mapping and device localization. Their approach treats the channel transfer function as an angle-delay matrix, where back-scattered channel impulse responses (CIRs) provide rich spatial information. Central to this study is a pose estimation algorithm operating in the radar coordinate system, incorporating an EKF for accurate localization. The measurements used for estimation include horizontal and vertical position shifts, as well as rotational deviations relative to previous poses. A notable innovation in the same study is the Fourier-Mellin-based algorithm, an FFT-based technique traditionally used for image registration by optimizing matches in the frequency domain \cite{Lotti2023_RadioSLAM_THz_6G}. The study applied this algorithm to consecutive radar frames in polar coordinates, treating them as `radio images` of the environment. By decoupling translation and rotation effects, the method improves relative pose estimation between two consecutive frames. The authors compared the Fourier-Mellin approach with the Laser Scan Matching algorithm, demonstrating superior performance in radio-based localization.

In \cite{Tong2018_FineLoc_WirelessLocalization}, the authors proposed FineLoc, a fine-grained, self-calibrating localization system that constructs radio maps using freely deployed Bluetooth low energy (BLE) nodes and crowd-sourced data. The study highlighted that existing systems struggle with inaccurate floor plan generation due to reliance on coarse-grained WiFi reference points. FineLoc addresses this limitation by leveraging BLE beacons as reference sources, incorporating dead-reckoning optimization and trace merging techniques to improve accuracy. The clustered trace segments are integrated into a skeleton of the floor plan, merging data from different relative coordinate systems into a unified global frame.

Finally, RF-based detection and tracking are subject to Doppler phase shifts, typically caused by the translational motion of a UE relative to a transmitter. However, additional phase shifts can arise due to mechanical vibrations experienced by the UE, particularly when mounted on a moving rigid body such as an autonomous ground vehicle. The study in \cite{Shastri2022_mmWave_Localization_Review} examined these disturbances, which result from variations in road profile and roughness, contributing to the effects of rotational motion. The authors modeled the Doppler phase shift using the transport theorem, expressed as \cite{Rao2006_Dynamics_ParticlesRigidBodies}:

\begin{equation}
f_D = \frac{2f}{c} (\mathbf{v} + \boldsymbol{\omega} \times \mathbf{p})^T \cdot \mathbf{n},
\end{equation}

where $f_D$ is the Doppler phase shift, $c$ the speed of light, $\mathbf{v}$ is the $3 \times 1$ translational velocity vector of the moving robot/vehicle, expressed in the moving frame, $\boldsymbol{\omega}$ is the $3 \times 1$ vector of angular velocities of the vehicle body with respect to the inertial frame and expressed in the body frame, $\mathbf{p}$ is the position vector describing the location of an arbitrary antenna point, and $\mathbf{n}$ is the radio transmitter/receiver LOS direction.

\subsection{WiFi Fingerprinting-based Localization}
RF sensors offer a robust solution for indoor SLAM applications, as they are resilient to dynamic lighting conditions and structured indoor environments while operating effectively even without a visual LOS. Traditional RF-based localization approaches rely on RSSI measurements, which require analytical modeling to characterize radio signal distribution over distance. However, due to the inherent complexity and variability of indoor signal propagation, the research community has increasingly adopted alternative methods that bypass the need for explicit analytical models \cite{Liu2020_CollaborativeSLAM_WiFiFingerprint}. One such alternative is WiFi fingerprinting, which represents a location by aggregating radio signals from multiple WiFi APs, offering greater robustness against signal distortions. WiFi radio wave equipment is widely deployed in indoor spaces, with many fixed anchors such as APs, making it an attractive alternative to other sensing devices. Compared to LiDAR sensors, WiFi-based SLAM benefits from significantly lower size, weight, power, and cost (SWaP-C) factors, making it a highly efficient choice \cite{Liu2019_CrowdsensingSLAM_PNT}. To achieve high positioning accuracy in large-scale environments, fine-grained radio maps are typically required. However, generating these maps is a time-intensive process, especially when relying on a single user for data collection. To address this, low-cost approaches such as crowdsensing, where multiple users contribute data, have been explored as a viable solution.

WiFi fingerprinting methods generally adopt learning-based localization techniques, where the similarity between a collected signal embedding and a recorded fingerprint is leveraged to infer the position of an agent. A wide range of ML techniques has been applied in this domain \cite{Roy2021_ML_IndoorLocalizationSurvey}, spanning from traditional heuristically derived, handcrafted feature-based models and feature selection algorithms to hierarchically self-evolving, DL models. Among these, DL approaches have demonstrated superior localization accuracy. Transfer learning has also been employed to mitigate the overhead of collecting vast amounts of fingerprints, facilitating the development of scalable and adaptive indoor localization systems (ILS). While DNNs such as DCNNs and recurrent neural networks (RNNs) have shown remarkable success in solving indoor localization challenges, their iterative learning processes can sometimes lead to slow convergence. To address this, extreme learning machines have been explored as an alternative, as they randomly generate hidden layer parameters, enabling extremely fast learning speeds and significantly reducing both training and testing times.

To additionally improve the accuracy of WiFi fingerprint-based SLAM and mitigate uncertainty, the authors of \cite{Liu2020_CollaborativeSLAM_WiFiFingerprint} proposed an approach that integrates WiFi fingerprinting with pedestrian dead reckoning (PDR) for crowdsensing-based SLAM in unknown indoor environments. Their key contributions are: (i) Uncertainty modeling for loop closures based on WiFi fingerprint similarity and short-term odometry measurements, (ii) Motion-based turning features’ incorporation to further reduce the uncertainty of radio fingerprint-based loop closures, enhancing overall accuracy and (ii) Empirical validation in a 9,000 m² University campus building, using two different PDR systems.

\begin{figure}[!t]
\centering
\includegraphics[width=\linewidth]{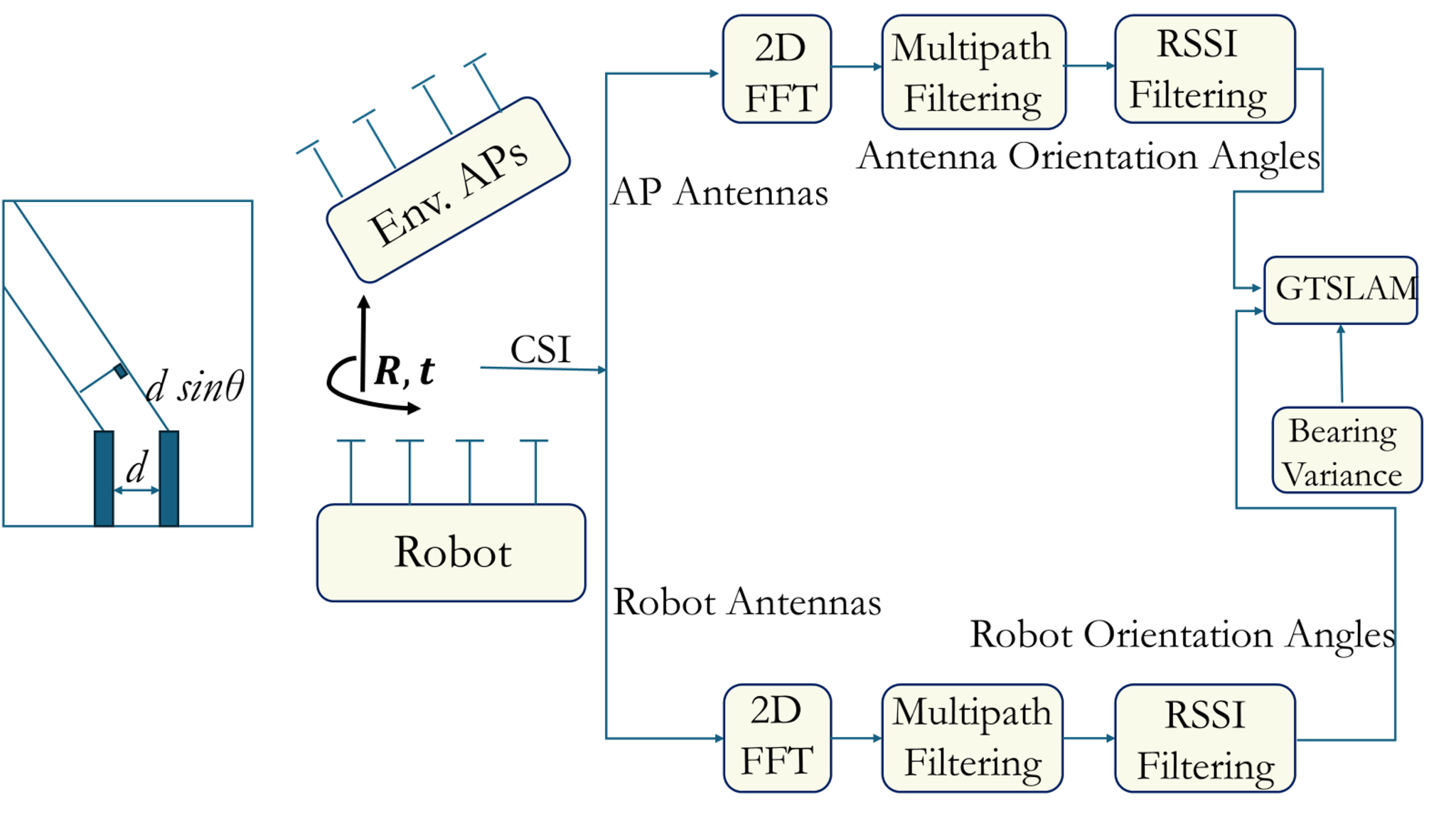}
\caption{Illustration of P²-SLAM incorporating wheel odometry, two-way bearings and a pose refinement via graph SLAM, based on \cite{Arun2022_P2SLAM_WiFiBearing}.}
\label{fig_12}
\end{figure}

Further demonstrating the potential of WiFi as a reliable and cost-effective SLAM sensor, the study in \cite{Arun2022_P2SLAM_WiFiBearing} presents P²-SLAM, a framework that eliminates the need for explicit loop closures while overcoming the limitations faced by RGB cameras and LiDAR sensors in challenging indoor environments. P²-SLAM integrates graph SLAM principles and consists of the following key components (Fig. \ref{fig_12}) \cite{Arun2022_P2SLAM_WiFiBearing}:

\begin{list}{}{}
\item{1. Wheel Odometry Integration - Utilizing odometry data to enhance trajectory estimation.}
\item{2. Two-Way Bearings - Estimating orientation angles from both the robot and the AP side to obtain the angle subtended by an AP at a given robot pose. This enables AP poses to be incorporated into the SLAM optimization process while accounting for prediction errors.}
\item{3. Pose Estimation Refinement - A prior pose estimate is incorporated into the factor graph-based optimization, further improving localization accuracy.}
\end{list}

\subsection{Device-based Sensing and Localization}

Device-based sensing plays a central role in ISAC, leveraging UE for environmental perception and localization. In bistatic sensing mode, a robust way to measure AoD and AoA angle measurements is crucial, as they are typically unknown. The system described in \cite{Yang2025_ISAC_PrototypeSystem} generates a 64 × 64 reference signal received power (RSRP) matrix after performing 64 × 64 beam scanning for a given BS-UE orientation pair. By combining multiple RSRP matrices from different BS-UE orientations, an RSRP image can be constructed with extended angular coverage, effectively representing the received signal in the beam space. This enables the transformation of AoA and AoD estimation from a traditional RF problem into an object detection problem in images. To address this, a DL-based framework is proposed, capable of estimating, tracking, and identifying mmWave channel multipath parameters while distinguishing between LOS and NLOS paths, effectively framing the problem as a universal computer vision task.

The study in \cite{Zeng2024_mmWaveSensing_LowRankCP} considers a mobile mmWave device sensing application, exploiting the small wavelength of mmWave signals to deploy large antenna arrays at both the transmitter and receiver. 5G mmWave communications are deemed to be inherently facilitating SLAM-enabled UE, capitalizing on larger bandwidths and antenna array augmentation as previously mentioned \cite{Yang2023_AngleBasedSLAM_5GmmWave}. Compared to sub-6 GHz communication, mmWave technology offers superior delay and angle measurements, making it well-suited for SLAM applications. These large arrays provide improved angular resolution, enabling precise environmental mapping. The transmitter scans the entire angular space, while the receiver captures echo signals reflected from scatterers in the environment. The locations of these scatterers are then extracted from the received signals, allowing the system to construct a spatial map of the surroundings. Extending the discussion to 6D pose estimation, the authors of \cite{Shen2024_6DPose_mmWaveMIMO} explore how UE can determine both position and orientation. They highlight the ability to integrate the huge antenna arrays into user devices while maintaining a small physical footprint, unlocking new possibilities for UE-based environmental awareness.

However, despite the discussed mmWave advantages, no prior work had successfully realized SLAM functionality within the 5G NR standard, primarily due to hardware and specification constraints. In their work, Yang et al. \cite{Yang2023_AngleBasedSLAM_5GmmWave} investigated how situational awareness can be achieved using 5G mmWave communication systems without modifying the transceiver architecture or the 5G NR standard. Their implementation features 28-GHz mmWave transceivers utilizing an OFDM-based 5G NR waveform with a 160-MHz channel bandwidth, following standard 5G NR beam management procedures. Additionally, they proposed an efficient successive cancellation-based angle extraction approach to estimate AoAs and AoDs from RSRP measurements. These angle measurements serve as the foundation for an angle-only SLAM algorithm, enabling simultaneous UE tracking and environmental feature mapping.

A critical challenge in realizing deep sensing-communication integration is leveraging off-the-shelf 5G communication hardware to perform different types of sensing while achieving mutual enhancement. In the context of 5G NR systems, active sensing is facilitated through uplink beam sweeping on the UE side, allowing for environmental observation. Simultaneously, passive sensing is achieved via downlink channel estimation, which extracts MPC-related information. The authors of \cite{Yang2022_HybridActivePassiveSLAM} presented the first hybrid SLAM mechanism that integrates active and passive sensing, introducing a common reflective surface feature to bridge the two approaches and enable information fusion. This fusion mechanism allows for: (i) physical anchor initialization based on MPC data from active sensing and (ii) probabilistic data association SLAM extension to achieve UE localization and continuously refine the physical anchor and target reflections through the subsequent passive sensing.

Considering the corner-case of user localization with a single-antenna receiver, the study in \cite{Fascista2021_SingleSnapshot_LocalizationMapping} proposed a novel approach for single-snapshot localization and mapping, aiming at addressing the aforementioned challenge. A joint maximum likelihood estimation problem was formulated, and its solution was formally derived. Given the computational burden of conducting a full-dimensional parameter search, the authors introduced an approximate maximum likelihood estimation method, which leverages the sparsity of mmWave channels to significantly reduce complexity while maintaining accuracy.

\subsection{Reconfigurable Intelligent Surfaces (RIS)-aided SLAM}

In the context of addressing the SLAM challenge in wireless environments, additional challenges posed by signal propagation, multipath effects, and dynamic radio environments need to be taken into account, as discussed in the previous section. One of the key challenges in wireless SLAM is dealing with uncontrollable radio environments where the structure and behavior of signals are unpredictable. A promising approach to mitigate these challenges is the MetaSLAM system proposed in \cite{Yang2022_MetaSLAM_RIS}, which incorporated multiple RISs. However, coordinating multiple RISs and optimizing their phase shifts, especially when their locations are unknown to the agent, poses significant difficulties. To address these challenges, the authors formulated a MetaSLAM optimization problem and developed a two-stage optimization algorithm based on genetic and particle filter algorithms to find a solution.

The MetaSLAM initialization phase consists of three sequential steps: communication, measurement acquisition, and localization and mapping \cite{Yang2022_MetaSLAM_RIS}.

\begin{list}{}{}
\item{1. Communication: During the first $\delta_O$ seconds, the communication step unfolds in three parts: (i) transmission of a starting signal from the transmitter to all RIS controllers to indicate the beginning of a new SLAM cycle denoted by k, (ii) each RIS controller responding with its unique identifier (ID) using time-division multiplexing (TDM) to ensure distinct signals, and (iii) the transmitter broadcasting a signal containing $U$ RIS ID values and the corresponding random vector $\boldsymbol{\Xi^k} = \{ \boldsymbol{\Xi_1^k}, \dots, \boldsymbol{\Xi_U^k} \}$ to the RIS controllers.} 
\item{2. Measurement Acquisition: Following the communication phase, the agent transmits a signal $s(t)$ to the environment while the receiver records the received signal $y_1 (t)$.}
\item{3. Localization and Mapping: Using $C$ estimated measurements $\boldsymbol{Z^k} = \{ \boldsymbol{z_1^k}, \dots, \boldsymbol{z_C^k} \}$, the initial agent position $\boldsymbol{p_a^0}$, initial velocity $\boldsymbol{v_a^0}$, and the initial localization and mapping estimation (LME) function $\boldsymbol{f^l}$, the updated agent position $\boldsymbol{p_a^k}$, velocity $\boldsymbol{v_a^k}$, and map $\boldsymbol{M^k}$ can be derived.}
\end{list}

In the optimization phase of the LME function, the position and map estimation step develops an RBPF that incorporates the following:

\begin{list}{}{}
\item{a. Agent Particle Transition - Updating position and velocity states using a first-order discrete-time state-space model with white noise.}
\item{b. MPC Weight Calculation - Assigning higher weights to MPCs with larger amplitudes due to their improved accuracy in time-of-arrival (ToA) and angle-of-arrival (AoA) estimation.}
\item{c. Landmark Update - Adjusting landmark location estimates using a Kalman filter correction step.}
\item{d. Weight Update of Agent Particles - Assigning importance factors to agent particles based on corresponding landmark associations.}
\item{e. Normalization and Resampling - Normalizing and resampling agent particles proportionally to their weights.}
\item{f. Agent and Landmark Update - Computing the agent's pose at a temporal cycle through a weighted sum of all agent particles.}
\item{g. Map Building - Estimating reflection point positions using dynamic MPCs and clustering reflections over different cycles to define reflectors.}
\end{list}

Beyond supporting SLAM through improved radio propagation or auxiliary measurements, recent research suggests that reconfigurable intelligent surfaces themselves can evolve into distributed nodes that directly provide SLAM services within programmable wireless environments. This perspective is motivated by the recognition that accurate spatial awareness is a fundamental requirement for reliable RIS operation, since errors in user localization or environmental understanding directly translate into suboptimal surface configuration and degraded system performance \cite{Tyrovolas2025OpticalAnchors}. As a result, RISs are increasingly being viewed as spatially anchored infrastructure elements capable of sensing, localization, and environment mapping, contributing persistent spatial information at the network level \cite{Zheng2024MultifunctionalRIS}. In this context, light-emitting RIS (LeRIS) architectures provide a particularly natural realization of RIS-native SLAM functionality, as the integration of optical anchors enables the surface to actively generate stable spatial references that can be exploited for localization and mapping independently of communication purposes \cite{Tyrovolas2025OpticalAnchors}. Specifically, by embedding optical emitters directly into RIS panels, these architectures allow localization and mapping capabilities to be shifted from mobile agents to the infrastructure, improving observability and robustness in dynamic environments \cite{Bozanis2024LeRIS}. More recently, VCSEL-based light-emitting RIS designs have demonstrated that a single surface can jointly support user localization, obstacle-aware mapping, and millimeter-wave communication by leveraging highly directional optical beams and low-complexity inference mechanisms \cite{Iqbal2025VCSELLeRIS}. Taken together, these developments point toward a paradigm in which RISs function as distributed SLAM service providers embedded within the environment, complementing agent-centric SLAM and enabling scalable and resilient networked autonomous systems.

\section{Challenges and Future Directions}
{\color{blue}
\subsection{Challenges}
}

The deployment of SLAM-enabled networked autonomous systems in real-world environments reveals a set of fundamental challenges that stem from the mismatch between idealized modeling assumptions and the complex dynamics of practical operation. 
\subsubsection{Illumination, Propagation, and Occlusion Effects}
Among these challenges, environment dynamics play a central role, since both the physical scene and the wireless propagation conditions evolve continuously over time. In {\color{blue}comparison with} classical SLAM formulations that assume static landmarks and slowly varying sensor parameters, real world scenarios are characterized by moving objects, time varying occlusions, and rapidly changing geometry. As such, motion {\color{blue}could directly affect} reliability and estimation consistency \cite{Bescos2018}. V-SLAM pipelines can experience unstable feature tracking and erroneous data association under illumination changes, motion blur, and partial occlusions, leading to map degradation and pose drift \cite{Scaramuzza2011,MurArtal2017}. Moreover, wireless-based localization and SLAM methods operating at mmWave and higher frequencies are highly sensitive to propagation variations, where blockage, Doppler shifts, and the intermittent appearance of dominant multipath components cause abrupt fluctuations in received signal strength, angle estimates, and CSI \cite{Rangan2014,Va2016}. As a result, when visual and wireless modalities are jointly considered, these dynamics further challenge system robustness, since the two sensing modalities respond differently to environmental changes and operate at different temporal resolutions, leading to cross modal inconsistencies and outdated representations \cite{Nishio2021}. As such, maintaining accurate localization and mapping in the presence of dynamic geometry and nonstationary sensing conditions remains an open challenge, motivating the development of SLAM frameworks that {\color{blue}account} for temporal variability and partial observability rather than relying on static environment assumptions.

\subsubsection{Latency Constraints}
As a consequence of operating in dynamic environments, SLAM-enabled networked autonomous systems are required to be responsive within strict time constraints, which elevates latency from an implementation concern to a fundamental challenge. Specifically, when scene geometry and wireless propagation conditions change within short time scales, delays in sensing, processing, communication, or inference can cause localization and mapping outputs to become misaligned with the current system state, thereby undermining their practical value. This challenge is particularly pronounced in networked SLAM architectures, where information must be exchanged among multiple agents or offloaded to edge or cloud infrastructures, introducing transmission and queuing delays that scale with network load and system size \cite{Satyanarayanan2017}. In vision-based SLAM, the increasing reliance on high resolution sensors and learning based perception modules further exacerbates latency, as computationally intensive front end processing and optimization stages may exceed real time constraints on embedded platforms \cite{cadena2016slamfuture}. Similarly, wireless-assisted SLAM approaches face latency limitations due to channel estimation and synchronization procedures, which restrict the temporal granularity at which CSI can be updated in high mobility scenarios \cite{Polese2018}. Moreover, in multimodal systems, heterogeneous sensing and processing pipelines introduce modality dependent delays, leading to asynchronous observations that challenge classical filtering and smoothing formulations \cite{Li2020}. Therefore, latency emerges as a system-level bottleneck, highlighting the need for SLAM frameworks that compensate for delayed measurements and seek a solution that adheres to a reasonable accuracy-latency trade off.

\subsubsection{Energy Efficiency}
Following the stringent latency requirements, energy efficiency emerges as a challenge, since achieving low latency in SLAM-enabled networked autonomous systems may come at the cost of increased computational effort. In more detail, in dynamic and networked environments, maintaining timely and accurate SLAM typically requires frequent sensing updates, intensive onboard processing, and continuous information exchange, all of which impose significant energy demands on mobile platforms with limited power budgets. This challenge is particularly important in vision-based SLAM systems, where high frame rate image acquisition and computationally demanding front end operations, including feature extraction optimization and deep learning based perception, can lead to substantial energy consumption on embedded hardware \cite{Chen2019}. Additionally, wireless-assisted SLAM and localization frameworks incur additional energy costs due to frequent CSI updates, especially at mmWave and higher frequencies where directional communication is required \cite{Alkhateeb2014}. Moreover, in networked SLAM architectures, energy consumption is further exacerbated by communication overhead associated with multi agent coordination and edge or cloud offloading, creating a fundamental trade off between local processing and transmission energy \cite{Wang2020}. Finally, in multimodal systems, jointly processing visual and wireless data streams can amplify energy requirements if fusion strategies are not carefully designed to exploit redundancy and conditional sensing opportunities. As a result, energy efficiency becomes a system level challenge that is closely related to latency and estimation performance, highlighting the need for adaptive SLAM frameworks that balance sensing computation and communication in an energy aware manner.

\subsubsection{Security and Privacy}
Finally, as SLAM-enabled networked autonomous systems increasingly rely on connectivity, distributed processing, and data sharing, attributes such as security and privacy emerge as additional challenges derived from the shift toward networked perception. In particular, the exchange of localization and mapping information among multiple agents or with edge and cloud infrastructures enlarges the system attack surface, and therefore exposes SLAM pipelines to cyber and physical threats that can directly affect estimation reliability and operational safety. Recent studies have {\color{blue}shown} that adversarial manipulation of sensory inputs or communication links can corrupt localization and mapping results, which, in turn, can lead to unsafe or misleading behavior in autonomous platforms \cite{Lippi2025,Ramalakshmi2024}. Vision-based SLAM systems are susceptible to visual spoofing and adversarial perturbations, whereas wireless-assisted SLAM frameworks may be affected by jamming, spoofing, or falsified CSI. The aforementioned can cause a degradation in both localization accuracy and communication performance \cite{SLACK2024}. Beyond integrity attacks, SLAM data inherently encodes sensitive information about environments and agent trajectories, including spatial layouts, mobility patterns, and potentially identifiable visual content, and thus raises significant privacy concerns when such data is stored, processed, or shared beyond the local platform \cite{Hataba2022}. These concerns are amplified in multimodal and networked settings. The fusion of visual and wireless data enables richer inference capabilities, at the cost, however, of increasing the potential for unintended information leakage and cross-domain attacks. Thus, security and privacy {\color{blue}can} be considered core challenges in SLAM-enabled networked autonomous systems, motivating the development of frameworks that incorporate robust estimation under adversarial conditions, secure communication mechanisms \cite{SHIELD}, and privacy-aware data representations as integral design components rather than auxiliary protections.
{\color{blue}
\subsection{Future Directions}
}
Based on the challenges discussed above, future research on SLAM-enabled networked autonomous systems is expected to further explore approaches that address environment dynamics, latency, energy efficiency, and security in an integrated manner. Since these challenges are tightly coupled at the system level, improving a single component is often sub-optimal, thereby motivating more holistic design principles that jointly consider sensing, communication, and computation. Recent advances in wireless communications, edge intelligence, and multimodal perception have led to increased interest in frameworks where SLAM and communication subsystems interact more closely, particularly in dynamic and networked environments. Rather than treating localization and mapping as independent of wireless operation, several works have investigated whether combining visual observations with radio measurements can improve robustness under fast-changing geometry and intermittent sensing conditions \cite{Nishio2021}. In vehicular and mobile robotic scenarios, for example, visual data acquired from low-cost sensors such as monocular cameras can be fused with wireless indicators including received signal strength or CSI to jointly estimate agent pose and motion-related states. By exploiting temporal correlations across multimodal data streams, such approaches may {\color{blue}suggest} short-term prediction of both mobility and wireless link dynamics, which is particularly relevant in environments affected by frequent occlusions and {\color{blue}NLOS} transitions. In addition, these predictive capabilities can support proactive communication functions, such as beam selection or handover decisions, while potentially reducing the need for frequent channel probing, and thus alleviating latency and energy constraints. Conversely, wireless measurements can provide complementary information when visual perception is degraded, offering an additional sensing modality that may help mitigate failures caused by adverse lighting or occlusions. Finally, recent studies on joint communication and sensing architectures suggest that such bidirectional interaction between perception and communication layers could form a foundation for SLAM-aware communication systems and communication-aware SLAM frameworks, especially in beyond-5G and 6G networked autonomy settings \cite{Wild2021}. Overall, these directions highlight a broader shift toward tightly integrated, predictive, and resource-aware SLAM architectures, where robustness, efficiency, and trustworthiness {\color{blue}addressed jointly may warrant further research in real-world environments}.

\section{Conclusions}
This survey has {\color{blue}examined elements of the emerging intersection} between SLAM and wireless communications, focusing on their growing interdependence in the era of mmWave/GHz communications, 5G, and beyond. The study highlighted how RF and visible light spectra camera modalities can serve as the basis for networked autonomous systems' to converge towards a unified localization, mapping, communication, and control framework. {\color{blue}Through a framework} integrating the foundations of SLAM in networked autonomous systems, SLAM hetero-modal techniques, mathematical methods, and wireless communication–enabled perception, the study was oriented towards the aforementioned convergence.

A key outcome of the information synthesis that was {\color{blue}undertaken} is the identification of the bidirectional relationship between wireless systems and SLAM. On one hand, wireless communications, through RF, WiFi, and mmWave signals, {\color{blue}have the potential for} accurate localization and NLOS situational awareness, especially under occluded visual conditions. On the other hand, computer vision and SLAM methodologies, including visual odometry, feature-based tracking for translational and rotational velocity retrieval, and feature-based landmark reconstruction, {\color{blue}providing} new tools for geometric modeling of communication channels, beam prediction, and mobility-aware network modeling. This reciprocity {\color{blue}shows} how perception and connectivity can evolve in parallel within a shared estimation and motion planning loop.

At an estimation-theoretical level, the survey revisited probabilistic modeling approaches in localization such as Bayesian filtering, belief-propagation-based state estimation whilst more sophisticated approaches such as vector-field SLAM built on a mathematical background that stood out from the crowd given an observed trend that equalizes SLAM with extended Kalman filtering. From the technological perspective of wireless communication, the discussion included DL–based localization, radio map construction, RIS-aided MetaSLAM, and multi-agent active estimation frameworks, {\color{blue}suggesting} deep convergence of communication, sensing, and control. The study also highlighted some emerging trends towards ISAC, where wireless infrastructures {\color{blue}can act} as both communication media and spatial sensors. To this end, passive SLAM was identified as a natural evolution of state observation in the context of networked autonomy, where perception is inherently maintained by rich visual and wireless signals without deliberately orienting the networked autonomous vehicle towards the direction of {\color{blue}increasing} information gain or estimation uncertainty gradient {\color{blue}decline}. Active SLAM, on the other hand, {\color{blue} can utilize} the system's control input vector to actively reduce uncertainty for a computer-vision-driven such as UAV cinematography or communications-driven such as beam tracking perception goal.

The findings of this study indicate that joint SLAM–wireless systems design {\color{blue}has the potential to} form a cornerstone for future networked autonomous systems, {\color{blue}facilitating} safe navigation, robust state feedback control, environment reconstruction, and real-time network optimization. Whilst early research {\color{blue}shows} promising results, fully integrated solutions {\color{blue}appear to be} in their infancy. Advancements in hetero-modal fusion, distributed estimation information exchange among autonomous agents, vision-language models in robotics, and edge-powered computation are {\color{blue}important for achieving} the next generation of visual perception-aware communications and communications-aware visual perception. Thus, the {\color{blue}intersection} of SLAM and wireless communications {\color{blue}forms} a foundational paradigm for 6G-enabled intelligent robotic and autonomous vehicle systems, {\color{blue}supporting} advanced reasoning on space and time through adaptive, unobstructed wireless links.

\bibliographystyle{IEEEtran}
\bibliography{references}

\begin{IEEEbiography}[{\includegraphics[width=1in,height=1.25in,clip,keepaspectratio]{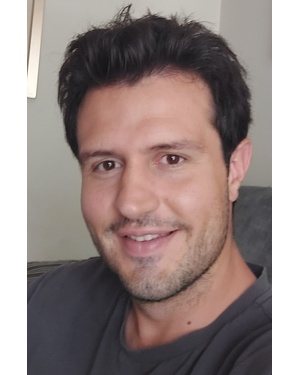}}]{Konstantinos Gounis}
received the Diploma (five years) from the Department of Production Engineering and Management, Democritus University of Thrace, Xanthi, Greece, in 2013, and completed his internship at the Department of Mechanical Engineering, Aristotle University of Thessaloniki, Thessaloniki, Greece, in 2014. He received an M.Sc. Degree from the Department of Informatics, Aristotle University of Thessaloniki, Greece, in 2024. He is a Ph.D. student at the School of Electrical and Computer Engineering, Aristotle University of Thessaloniki. His research interests are focused around Vision-aided Sensing and Communications and Autonomous Systems. 
\end{IEEEbiography}

\begin{IEEEbiography}[{\includegraphics[width=1in,height=1.25in,clip,keepaspectratio]{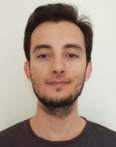}}]{Sotiris A. Tegos}
(Senior Member, IEEE) received the Diploma (five years) and Ph.D. degrees from the Department of Electrical and Computer Engineering, Aristotle University of Thessaloniki, Thessaloniki, Greece, in 2017 and 2022, respectively. Since 2022, he has been a Postdoctoral Fellow with the Wireless Communications and Information Processing Group, Aristotle University of Thessaloniki. In 2018, he was a Visitor Researcher with the Department of Electrical and Computer Engineering, Khalifa University, Abu Dhabi, UAE. His current research interests include multiple access in wireless communications, wireless power transfer, and optical wireless communications. He received the Best Paper Award in 2023 Photonics Global Conference and in 2025 IEEE Wireless Communications and Networking Conference. He serves as an Editor for \emph{\textsc{IEEE Transactions on Communications}} and \emph{\textsc{IEEE Communication Letters}}. He was an Exemplary Reviewer of \emph{\textsc{IEEE Wireless Communications Letters}} in in 2019, 2022, and 2023 (top 3\% of reviewers) and an Exemplary Editor of \emph{\textsc{IEEE Communications Letters}} in 2024 and 2025.
\end{IEEEbiography}

\begin{IEEEbiography}[{\includegraphics[width=1in,height=1.25in,clip,keepaspectratio]{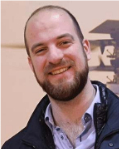}}]{Dimitrios Tyrovolas}
(Member, IEEE) received the Diploma degree in electrical and computer engineering from the University of Patras, Greece, in 2020, and the Ph.D. degree from the Department of Electrical and Computer Engineering, Aristotle University of Thessaloniki, Thessaloniki, Greece, in
2024, where he has been a Postdoctoral Fellow with the Wireless Communications and Information Processing Group since 2024. He is also a Research Assistant with Dienekes SI IKE, Heraklion, Greece. His current research interests include reconfigurable networking, reconfigurable intelligent surfaces, pinching-antenna systems, and UAV communications. He serves as an Editor for \emph{\textsc{IEEE Transactions on Communications}} and \emph{\textsc{IEEE Communication Letters}}. He was an Exemplary Reviewer of \emph{\textsc{IEEE Wireless Communications Letters}}, in 2021, and \emph{\textsc{IEEE Communication Letters}}, in 2022 (top 3\% of reviewers) and an Exemplary Editor of \emph{\textsc{IEEE Communications Letters}} in 2025.
\end{IEEEbiography}

\begin{IEEEbiography}[{\includegraphics[width=1in,height=1.25in,clip,keepaspectratio]{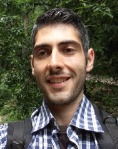}}]{Panagiotis D. Diamantoulakis}
(Senior Member, IEEE) received the Diploma and Ph.D. degrees from the Department of Electrical and Computer Engineering, Aristotle University of Thessaloniki (AUTH), Thessaloniki, Greece, in 2012 and 2017, respectively, where he was a Postdoctoral Fellow with the Wireless Communications and Information Processing Group from 2017 to 2024. Since 2021, he has been a Visiting Assistant Professor with the Key Lab of Information Coding and Transmission, Southwest Jiaotong University, Chengdu, China. In May 2024, he joined the faculty with AUTH, where he is currently an Assistant Professor with the Department of Electrical and Computer Engineering, Aristotle University of Thessaloniki, Thessaloniki, Greece. His research interests include optimization theory and applications in wireless networks, optical wireless communications, and goal-oriented communications. He serves as an Editor for the \emph{\textsc{IEEE Open Journal of the Communications Society}}. From 2018 to 2023, he was an Editor of \emph{\textsc{IEEE Wireless Communications Letters}}, in which he was an Exemplary Editor in 2020. He was an Exemplary Reviewer of \emph{\textsc{IEEE Communications Letters}} in 2014 and of \emph{\textsc{IEEE Transactions on Communications}} in 2017 and 2019 (top 3\% of reviewers).
\end{IEEEbiography}

\begin{IEEEbiography}[{\includegraphics[width=1in,height=1.25in,clip,keepaspectratio]{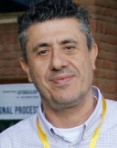}}]{George K. Karagiannidis}
(Fellow, IEEE) received the Ph.D. degree in telecommunications engineering from the Department of Electrical Engineering, University of Patras, Patras, Greece, in 1998. He is currently a Professor with the Department of Electrical and Computer Engineering, Aristotle University of Thessaloniki, Thessaloniki, Greece, and the Head of the Wireless Communications and Information Processing Group. His research interests are in the areas of wireless communication systems and networks, signal processing, optical wireless communications, wireless power transfer, and signal processing for biomedical engineering. He has recently received three prestigious awards: the 2021 IEEE ComSoc RCC Technical Recognition Award, the 2018 IEEE ComSoc SPCE Technical Recognition Award, and the 2022 Humboldt Research Award from the Alexander von Humboldt Foundation. He is one of the Highly Cited Authors across all areas of electrical engineering, recognized by Clarivate Analytics as a Web of Science Highly Cited Researcher from 2015 to 2024. He is currently the Editor-in-Chief of \emph{\textsc{IEEE Transactions on Communications}}. He was the Editor-in-Chief of \emph{\textsc{IEEE Communications Letters}}.
\end{IEEEbiography}

\vfill
\end{document}